%% file: acl_latex.tex
\documentclass[11pt]{article}

\usepackage[final]{acl}
\usepackage{times}
\usepackage{latexsym}

\usepackage[T1]{fontenc}

\usepackage[utf8]{inputenc}
\usepackage{microtype}
\usepackage{inconsolata}
\usepackage{graphicx}
\graphicspath{{../}{../figures/}}

\usepackage{CJKutf8}
\usepackage{amsmath}
\usepackage{amssymb}
\usepackage{dsfont}
\usepackage{kotex}
\usepackage{booktabs}
\usepackage{tabularx}
\usepackage{multirow}
\usepackage{multicol}
\usepackage{enumitem}
\usepackage{subcaption}
\usepackage[table]{xcolor}

\newcommand{\ours}{DLN}
\newcommand{\lape}{LAPE}
\newcommand{\ternarylsn}{LSN}
\newcommand{\ternarylrn}{LRN}
\newcommand{\selectedj}{j^\star}

\title{Distribution-aware Language Neuron Identification\\in Multilingual Large Language Models}

\author{
    Minjun Kim\textsuperscript{1}\hspace{1cm} Inho Won\textsuperscript{1}\hspace{1cm} Junghun Yuk\textsuperscript{1}\hspace{1cm} 
    Dongyeon Kim\textsuperscript{1} \\ \bf Jihyo Kim\textsuperscript{2}$^{\dagger}$\hspace{1cm} KyungTae Lim\textsuperscript{1}$^{\dagger}$ \\
    \textsuperscript{1}KAIST\hspace{4mm} \textsuperscript{2} KAIST InnoCORE PRISM-AI Center \\
    {\{mjkmain, ktlim\}@kaist.ac.kr} \\
}

\begin{document}
\maketitle
{\renewcommand{\thefootnote}{\fnsymbol{footnote}}%
 \footnotetext[2]{Corresponding Authors}}

\begin{abstract}
Multilingual large language models (mLLMs) contain a small fraction of feed-forward neurons that are sensitive to particular languages, commonly termed language-specific neurons. Existing work measures language specificity using the entropy of each neuron's language-wise probabilities of being active, where a neuron is considered active when its activation value is positive.
However, this approach may not fully capture the multilingual nature of mLLMs, where language representations are distributional and mutually related.
We propose \textit{Distribution-aware Language Neuron} selection, which leverages pairwise relationships between per-language activation distributions over the full activation range, including negative values. Specifically, we quantify each neuron's language specificity by clustering languages using pairwise overlap coefficients between their activation distributions.
Across two mLLMs and two held-out corpora, our identifier more effectively isolates language-specific causal effects, yielding up to 4.9$\times$ higher on-target language damage per neuron while preserving off-target language performance.

\end{abstract}

\input{sections/01_introduction}
\input{sections/02_related_work}

\input{sections/03_method}
\input{sections/04_experiments}
\input{sections/05_analysis}

\input{sections/06_conclusion}
\input{sections/98_limitations}

\section*{Use of AI Assistants}
We used an AI assistant (Gemini) to polish the writing of this paper, limited to grammar correction and sentence-level wording suggestions. All suggestions were reviewed and revised by the authors.

\section*{Acknowledgements}
This work was supported by the Institute of Information \& Communications Technology Planning \& Evaluation (IITP) grant funded by the Korea government (MSIT) (No.RS2024-00456709, A Development of Self-Evolving Deepfake Detection Technology to Prevent the Socially Malicious Use of Generative AI), and by the InnoCORE program of the Ministry of Science and ICT(N10260002). This work utilized GPU resources from the ``Advanced GPU Utilization Support Program'' funded by MSIT, Republic of Korea (awarded to KyungTae Lim).

\bibliography{custom}

\clearpage
\appendix
\input{sections/99.appendix}

\end{document}

%% file: sections/01_introduction.tex
\section{Introduction}

Multilingual large language models (mLLMs) exhibit strong capabilities across dozens of languages within a single shared parameter space~\citep{workshop2022bloom,yang2025qwen3}. However, how these models internally organize multilingual representations remains an open question.

\begin{figure}[t]
\centering
\includegraphics[width=\linewidth]{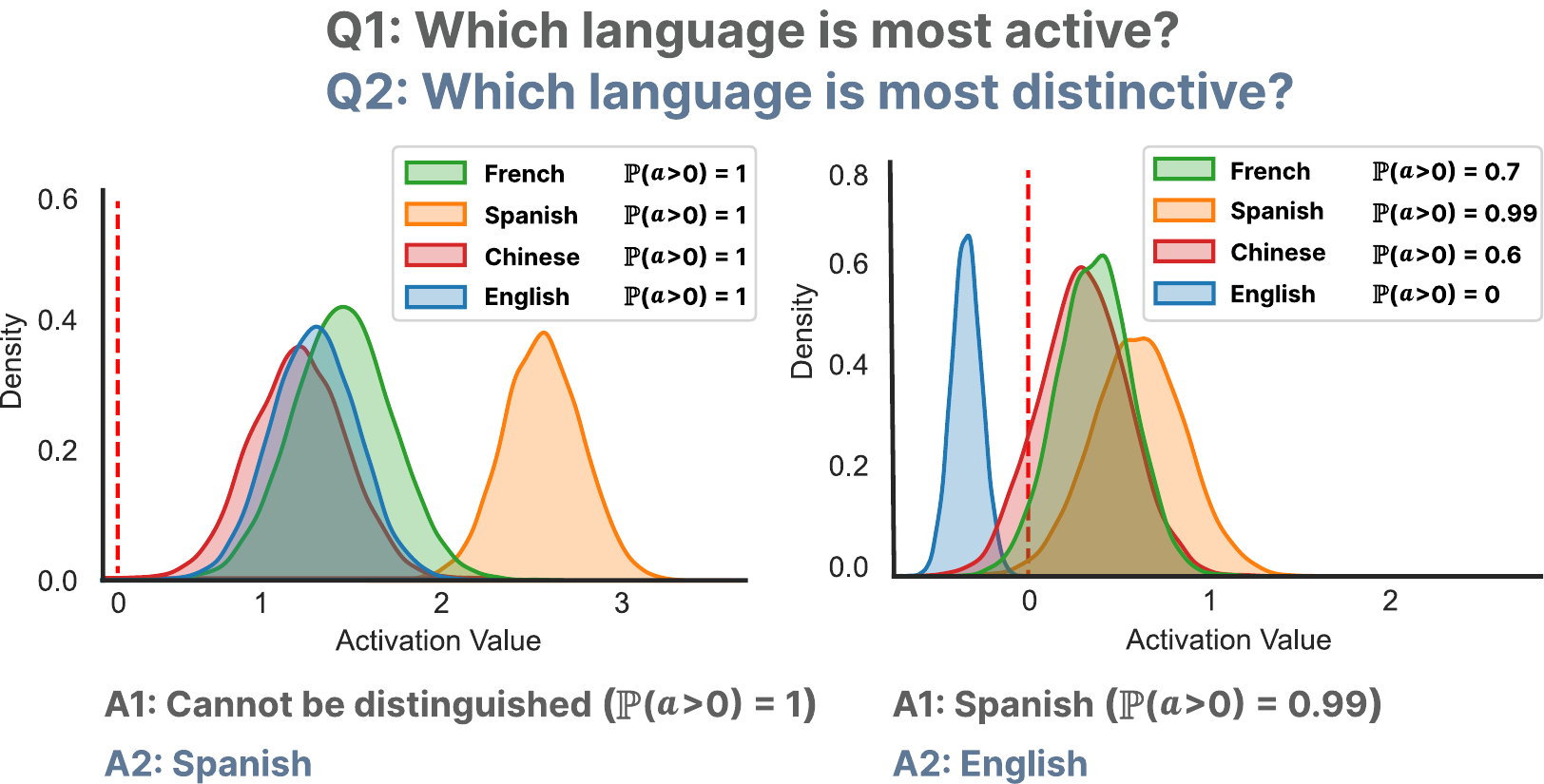}
\caption{Identifying structurally distinctive language distributions matters more than identifying the most frequently active language. Here, $a$ denotes the activation value. A distributional view reveals distinctions that activation probabilities alone miss.}

\label{fig:figure1}
\vspace{-0.16in}
\end{figure}

One empirical approach to this question is neuron-level interpretability based on activation values. Previous studies identify language-specific neurons~(LSNs) as neurons that exhibit preferential activation for particular languages, based on activation profiles that distinguish one language from the others~\citep{tang2024language, kojima-etal-2024-multilingual, zhang2026does}.
They consider neurons with positive activation values to be activated for each language
and quantify language specificity using the entropy of per-language activation probabilities. 
In this view, neurons exhibiting language specificity can be conceptualized as
``neurons that activate predominantly for a single language.'' These identified neurons provide a useful analytic tool for studying the mechanisms of cross-lingual transfer, intervening on language generation, and probing multilingual knowledge organization~\citep{rahmanisa-etal-2025-unveiling,mondal-etal-2025-language,deng-etal-2025-unveiling}.

However, previous LSN identification methods may not fully reflect the multilingual nature of mLLMs. Since mLLMs are trained on multiple languages under a shared parameter space, language representations are inherently distributional and mutually related. Binarizing continuous activations and treating them as separate per-language distributions introduces two critical limitations: 
first, it discards informative signals from negative activations since neurons with negative activation values are treated as inactive; second, because activation probabilities are computed from a binary state separately for each language, they do not preserve the unique shape of per-language activation distributions and the relational structure among languages. 

Figure~\ref{fig:figure1} provides an example where the previous activation probability methods are insufficient to capture the language specificity. As shown in the left plot, even when all activation values are positive, previous methods may fail to identify this neuron as language-specific because the entropy remains high despite distinct activation distributions.
Furthermore, the right plot shows a scenario in which focusing solely on activation probabilities may fail to capture the structural distinctiveness: in this example, English forms an isolated distribution in a region that previous methods would treat as inactive. These structural blind spots raise a fundamental question: given the multilingual nature of mLLMs, how can we identify language neurons by leveraging the activation distribution and its relational structure across languages while avoiding potential information loss? %

To address this question, we propose \textit{Distribution-aware Language Neuron} (DLN) selection, which estimates neuron-level language specificity from pairwise overlap coefficients between per-language activation distributions. 
By modeling the full range of activation values and clustering languages based on their pairwise overlap coefficient, DLN captures the relational structure among activation distributions. This formulation naturally identifies two types of language neurons, (1) \emph{single-language neurons} (SLNs), specific to one language, and (2) \emph{multi-language neurons} (MLNs), specific to multiple languages. Consequently, rather than focusing solely on neurons that activate predominantly for a single language, our method reframes the neurons to be identified as ``neurons whose activation distributions differ across languages.''


Through evaluations on Llama-3.1-8B \citep{grattafiori2024llama} and SmolLM3-3B-Base \citep{bakouch2025smollm3} over two held-out corpora, with additional validation on three further models in the appendix, we show that DLN more effectively isolates language-specific causal effects than activation-probability baselines. Ablating DLN-selected SLNs yields up to 4.9$\times$ higher on-target language damage per neuron while preserving off-target performance. DLN also captures MLNs, uncovering shared structures such as a Chinese--Japanese coalition that yields 7.12$\times$ greater within-cluster damage while using 2.1$\times$ fewer neurons. By modeling activation densities, DLN identifies neurons operating in the negative activation regime, including English SLNs that are invisible to positive-rate metrics yet act as functional triggers for target-language control. Finally, DLN remains effective beyond the standard gate-activation site, maintaining high selectivity at the gate-up product site where activation-probability baselines lose much of their effectiveness. Our contributions are as follows:
\begin{itemize}[leftmargin=*]
\item \textbf{A distribution-aware language neuron identification framework.} By formulating neuron selection as a clustering problem using overlap coefficients of continuous activation densities, our method effectively leverages the distributional and inter-language relational structures reflecting the nature of mLLMs.
\item \textbf{Effectiveness across SLNs and MLNs.} We evaluate the effectiveness of the neurons identified by our method for two types of language neurons. SLNs significantly outperform the baselines in isolating language-specific causal effects, while MLNs identify structurally coupled language coalitions. 
\item \textbf{What full activation distributions reveal.} Analysis results demonstrate that our method achieves its design objectives: (1) capturing the importance of neurons identified in the negative activation regime, and (2) enabling flexible intervention locations by considering the full range of activation values.


\end{itemize}

%% file: sections/02_related_work.tex
\section{Related Work}

mLLMs share parameters across many languages while preserving language identity in generation~\citep{conneau-etal-2020-unsupervised,grattafiori2024llama,yang2025qwen3}. Prior work suggests that languages occupy partially overlapping representational subspaces~\citep{pires-etal-2019-multilingual,wu-dredze-2019-beto} and that decoder-only mLLMs may internally rely on English-centric pivot representations~\citep{wendler-etal-2024-llamas,schut2025do}. These findings motivate a mechanistic question: how are language-specific and language-shared computations implemented inside a single mLLM?

\begin{figure*}[!t]
\centering
\includegraphics[width=\textwidth]{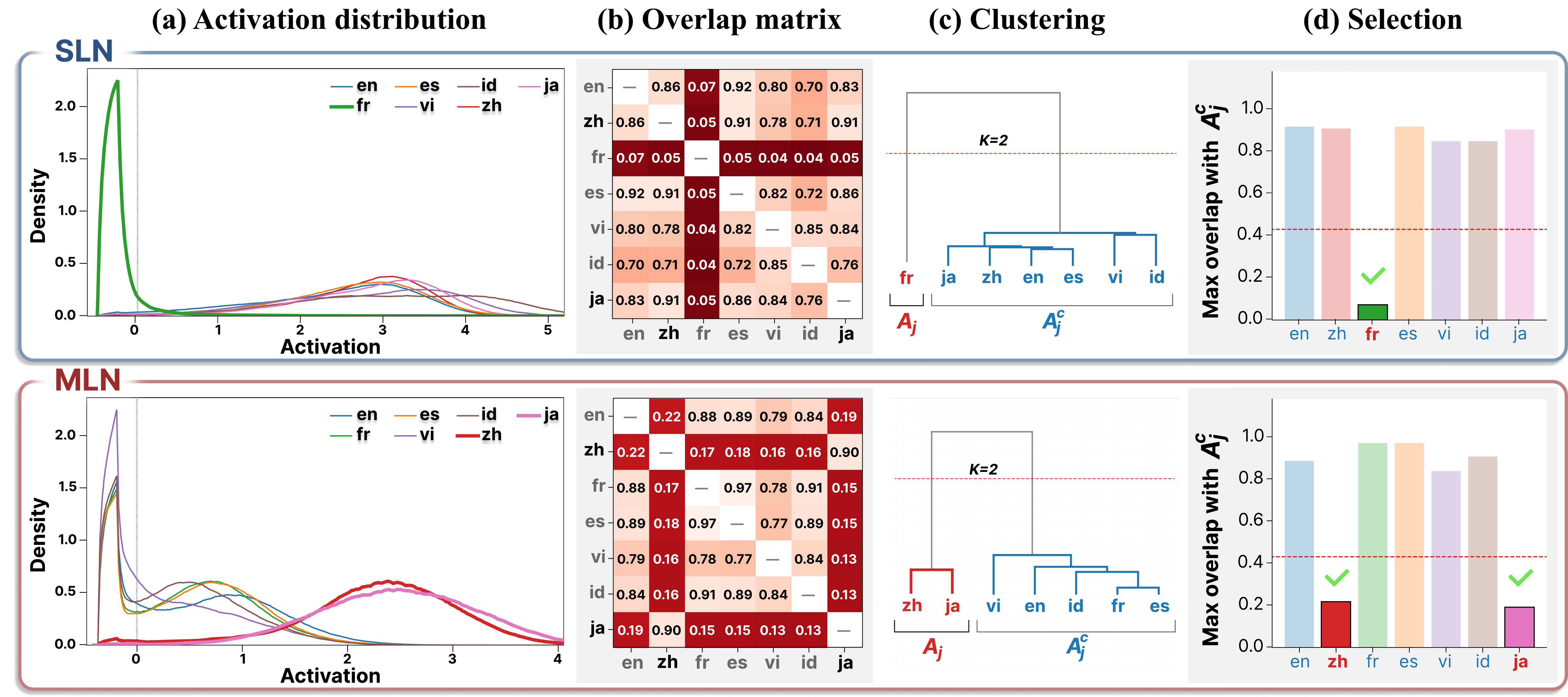}
\caption{Overall framework of the proposed method.}
\label{fig:framework}
\vspace{-0.1in}
\end{figure*}

Neuron-level analyses provide one way to study this question. Prior work shows that individual neurons can encode factual, syntactic, semantic, or task-relevant information in Transformer models~\citep{geva-etal-2021-transformer,dai-etal-2022-knowledge,stanczak-etal-2022-neurons}. Extending this line to multilingual models, \citet{kojima-etal-2024-multilingual} identify language-specific neurons~(LSNs) by measuring how precisely each neuron distinguishes a target language from others. \citet{tang2024language} introduce language activation probability entropy~(LAPE), which measures language specificity from the entropy of per-language activation probabilities, treating neurons with positive activation values as active. Based on LAPE, \citet{zhang2026does} categorize neurons as LSNs, language-related neurons~(LRNs), and general neurons based on how many languages exceed a predefined activation-probability threshold.\footnote{LSNs and LRNs are closely related to our notions of single-language neurons~(SLNs) and multi-language neurons~(MLNs), respectively. We use SLNs and MLNs to emphasize whether the identified neurons are associated with a single language or shared across multiple languages.}

However, these methods mainly rely on binary activation statistics, i.e., whether and how often a neuron is active for each language. This abstraction can miss distributional distinctions: languages with similar activation probabilities may still induce different activation-value distributions, and vice versa. In contrast, we analyze the full per-language activation distribution, enabling the identification of both single-language and multi-language neurons from distributional activation patterns.

%% file: sections/03_method.tex
\section{Distribution-aware Language Neurons}\label{sec:distribution-aware}


In this section, we introduce \emph{distribution-aware language neuron (DLN)} identification, which leverages the full range of per-language activation distributions and quantifies their relationships using pairwise overlap coefficients. Previous methods reduce each per-language activation distribution to a binary activation statistic based on the sign of activation values, although negative activations may contain useful information for determining language specificity. Moreover, multiple languages can have nearly identical binary activation statistics while occupying clearly separable activation regimes. Such neurons may still be language-specific, but entropy-based methods can fail to identify them. Figure~\ref{fig:framework} illustrates the overall framework of our method.

\subsection{Distribution Overlap Coefficient Matrix}

We study autoregressive Transformer language models with Gated Linear Unit (GLU) feed-forward blocks~\citep{shazeer2020glu}. We index each multilayer perceptron (MLP) neuron by the pair $j = (\ell, n)$, where $\ell$ is its layer and $n$ is its position within the layer's intermediate dimension $d_\mathrm{ffn}$. Let $d$ denote the residual-stream hidden dimension. Given the gate projection $W_\mathrm{gate}^\ell \in \mathbb{R}^{d_\mathrm{ffn}\times d}$, the activation of neuron $j$ on input $x \in \mathbb{R}^d$ is
\begin{equation*}\label{eq:activation}
a_j(x) \;=\; \bigl[\sigma(W_\mathrm{gate}^\ell x)\bigr]_n,
\end{equation*}
where $\sigma(\cdot)$ is the activation function (e.g., SiLU).


We determine each neuron’s language specificity by measuring the overlap coefficients, the shared areas, between per-language activation distributions. Let $\mathcal{L}$ denote the set of languages, and $p_j^k(v)$ denote the continuous probability density function of the activation value $v = a_j(x)$ for neuron $j$ under
language $k\in\mathcal{L}$. The pairwise overlap coefficient matrix 
$\mathbf{S}_j \in [0,1]^{|\mathcal{L}| \times |\mathcal{L}|}$ is calculated as:
\begin{equation*}
\mathbf{S}_j[k, k'] = \int \min\left(p_j^k(v),\, p_j^{k'}(v)\right) dv.
\end{equation*}
This summarizes the relationship among $\{p_j^k(v)\}.$

The overlap coefficient is one of several bounded, shape-sensitive statistics for
comparing per-language distributions, and we adopt it as the most intuitive and
computationally cheapest choice. Appendix~\ref{app:metric-linkage} shows that alternatives such as Jensen-Shannon, squared Hellinger, and Kolmogorov-Smirnov yield near-identical neuron sets, whereas scale-bearing metrics (Wasserstein-1, energy distance~\citep{SZEKELY20131249}) diverge.

\subsection{Cluster-based Neuron Classification}\label{subsec:cluster}
To distinguish languages with distinct activation patterns, we partition $\mathcal{L}$ into two clusters, $(A_j, A_j^c)$, by minimizing the maximum between-cluster overlap coefficient, $\max_{k \in A_j,\, k' \in A_j^c}\, \mathbf{S}_j[k, k']$, over all non-trivial bipartitions. This procedure is equivalent to single-linkage clustering~\citep{johnson1967hierarchical} with the number of clusters $K=2$. This is a selection decision rather than a bimodality assumption, as $\mathbf{S}_j$ retains all pairwise relations and a neuron with three or more separated groups is still detected through its most-separated bipartition. Appendix~\ref{app:metric-linkage} validates this design together with alternative linkage methods.

We define $A_j$ to be the smaller cluster and designate it as the neuron's candidate language set. This follows from the assumption of our method design that languages with distinct activation patterns are those most strongly associated with the neuron. The larger cluster $A_j^c$ can be regarded as a reference group of languages with relatively similar activation behavior.

\subsection{Neuron Selection Criterion}

After obtaining $A_j$ and $A_j^c$, we select neuron $j$ as a language neuron only if all between-cluster overlap coefficients do not exceed a threshold $\tau$:
\begin{equation*}
\mathcal{N} = \left\{ j: \max\limits_{k\in A_j, k' \in A_j^c} \mathbf{S}_j[k, k'] \le \tau\right\}.
\end{equation*}
Otherwise, if any between-cluster pair has an overlap coefficient greater than $\tau$, the neuron is not selected.

We denote each selected language neuron as ${\selectedj} \in \mathcal{N}$ and categorize it according to the size of its language set $A_{\selectedj}$ as follows:
\begin{itemize}
\item $|A_{\selectedj}| = 1$: \textbf{single-language neuron (SLN)},
\item $|A_{\selectedj}|>1$: \textbf{multi-language neuron (MLN)}.
\end{itemize}

Because overlap coefficients vary across models, we set the threshold $\tau$ to the bottom $P$-th percentile of overlap coefficients computed over all neurons.



%% file: sections/04_experiments.tex
\section{Experiments}
\label{sec:experiments}

\subsection{Experimental Settings}
\label{subsec:experiment_settings}

\paragraph{Baselines.}\label{subsec:baselines}
We employ two language neuron identification methods using the entropy of activation probabilities, LAPE~\citep{tang2024language} and \citet{zhang2026does}, as our baselines. Specifically, LAPE identifies LSNs by their defined score computed as:
\begin{equation*}
\mathrm{LAPE}_j \;=\; -\sum_{k \in \mathcal{L}} \tilde{r}_j^k \log \tilde{r}_j^k,
\end{equation*}
where $\tilde{r}_j^k$ is L1 normalized $r_j^k=\mathbb{P}\bigl(a_j(x) > 0|k\bigr)$ across $k$.

\citet{zhang2026does} additionally incorporate the maximum activation probability, $\max_{k \in \mathcal{L}} r_j^k$, to quantify the strongest degree to which neuron $j$ is active across languages to improve language neuron selection. Accordingly, they revise the LAPE score as follows:
\begin{equation*}
\mathrm{score}_j \;=\; -\sum_{k \in \mathcal{L}} \tilde r_j^k \log \tilde r_j^k \;-\; \lambda \max_{k \in \mathcal{L}} r_j^k,
\end{equation*}
where $\lambda$ is a balancing coefficient. The $\lambda$ for each evaluated model can be found in Table~\ref{app:lambda} in Appendix~\ref{app:implementation}. In addition, they distinguish LRNs from LSNs as neurons involved in multiple languages.\footnote{Hereafter, for clarity, we refer to LSNs identified by \citet{tang2024language} as LAPE, and to LSNs and LRNs identified by \citet{zhang2026does} as LSN and LRN, respectively.} Specifically, neuron $j$ is considered an LRN if the number of languages for which $r_j^k$ exceeds the predefined threshold is greater than $1$.

\paragraph{Models.}
We study two pretrained base models with Gated Linear Unit feed-forward design \citep{shazeer2020glu} and SiLU activation: Llama-3.1-8B~\citep{grattafiori2024llama} as our primary model, with SmolLM3-3B-Base~\citep{bakouch2025smollm3}.

\paragraph{Corpora and languages.}
For language neuron selection, we sample $100$k tokens per language from \texttt{wikipedia-20231101}. For evaluation, we use FLORES+ devtest (\citealp{nllb-24}; FLORES) and a held-out Wikipedia split (\texttt{wikipedia-20230601}; WIKI), up to $1\text{M}$ tokens per language, disjoint from the calibration source.

Following \citet{tang2024language}, we use seven languages spanning four scripts (Latin, Han, Kanji, and Kana) and five language families (Indo-European, Sino-Tibetan, Austroasiatic, Austronesian, and Japonic): English (\texttt{en}), Chinese (\texttt{zh}), French (\texttt{fr}), Spanish (\texttt{es}), Vietnamese (\texttt{vi}), Indonesian (\texttt{id}), and Japanese (\texttt{ja}).


\paragraph{Density approximation.}
To compute the continuous overlap integral in practice, we approximate the density $p_j^k(v)$ using a histogram-based approach with $400$ inner bins. The bins are defined on a per-neuron adaptive range $[\mu_j - 3\sigma_j,\,\mu_j + 3\sigma_j]$, where $\mu_j$ and $\sigma_j$ are pooled estimates over the seven languages.

\paragraph{Neuron intervention.}
For neuron $\selectedj$ identified for target $k$, we replace its $a_{\selectedj}$ with the mean activation value over the non-target languages. This mean-patching intervention aligns with our method: we treat $p_j^k$ as primitive and therefore call for replacing \(a_{\selectedj}\) with a non-target reference value estimated from the other languages, whereas the baseline methods operationalize language specificity through the binary event \(a_{\selectedj}(x)>0\), for which zero serves as a natural counterfactual.  Mean-patching realizes this intervention at the first-moment level while keeping the replacement within the neuron’s typical activation range~\citep{heimersheim2024use}. We verified that mean-patching does not adversely affect the baselines compared with zero-ablation, as detailed in Figures~\ref{fig:appendix-zeroout-llama}, \ref{fig:appendix-zeroout-smollm} in Appendix~\ref{app:zero-ablation}.

\paragraph{Selectivity score.}
Let $\mathcal{A}$ denote the collection of all target language clusters under evaluation. For $A \in \mathcal{A}$, let $A^c$ denote its complement $\mathcal{L}\setminus A$. Mean-patching its neuron subset $\mathcal{N}_A$ induces a negative log-likelihood (NLL) increase $\Delta\mathrm{NLL}_k(\mathcal{N}_A)$ on evaluation language $k$. The selectivity of neuron set $\mathcal{N}$ is:
\begin{equation*}
    \mathrm{Sel}(\mathcal{N}) :=
    \frac{\sum_{A\in \mathcal{A}}\!\left[\frac{1}{|A|}\sum_{k \in A} \Delta\mathrm{NLL}_k(\mathcal{N}_A)\right]}
         {\sum_{A\in \mathcal{A}}\!\left[\frac{1}{|A^c|}\sum_{k' \in A^c} \Delta\mathrm{NLL}_{k'}(\mathcal{N}_A)\right]}.
\end{equation*}
For instance, in Figure~\ref{fig:q1_heatmap}, $\mathrm{Sel}(\mathcal{N})$ is computed as the ratio between the diagonal sum and the off-diagonal sum. Higher $\mathrm{Sel}(\mathcal{N})$ means the intervention concentrates damage on the target languages $A$ and spares the off-target languages $A^c$.

\subsection{Assessing the Neuron Selectivity}\label{subsec:q1_identification}
We evaluate the effectiveness of our method from two perspectives: (1) for SLNs, whether it yields more language-selective per-language interventions than the baselines; and (2) for MLNs, whether our clustering-based framework recovers cluster-internal intervention effects missed by single-language identifiers and concentrates these effects more strongly per neuron than LRN. To examine these two perspectives, we analyze changes in NLL across languages after intervening on neurons associated with the target language.

\begin{figure}[t]
\centering
\includegraphics[width=\linewidth]{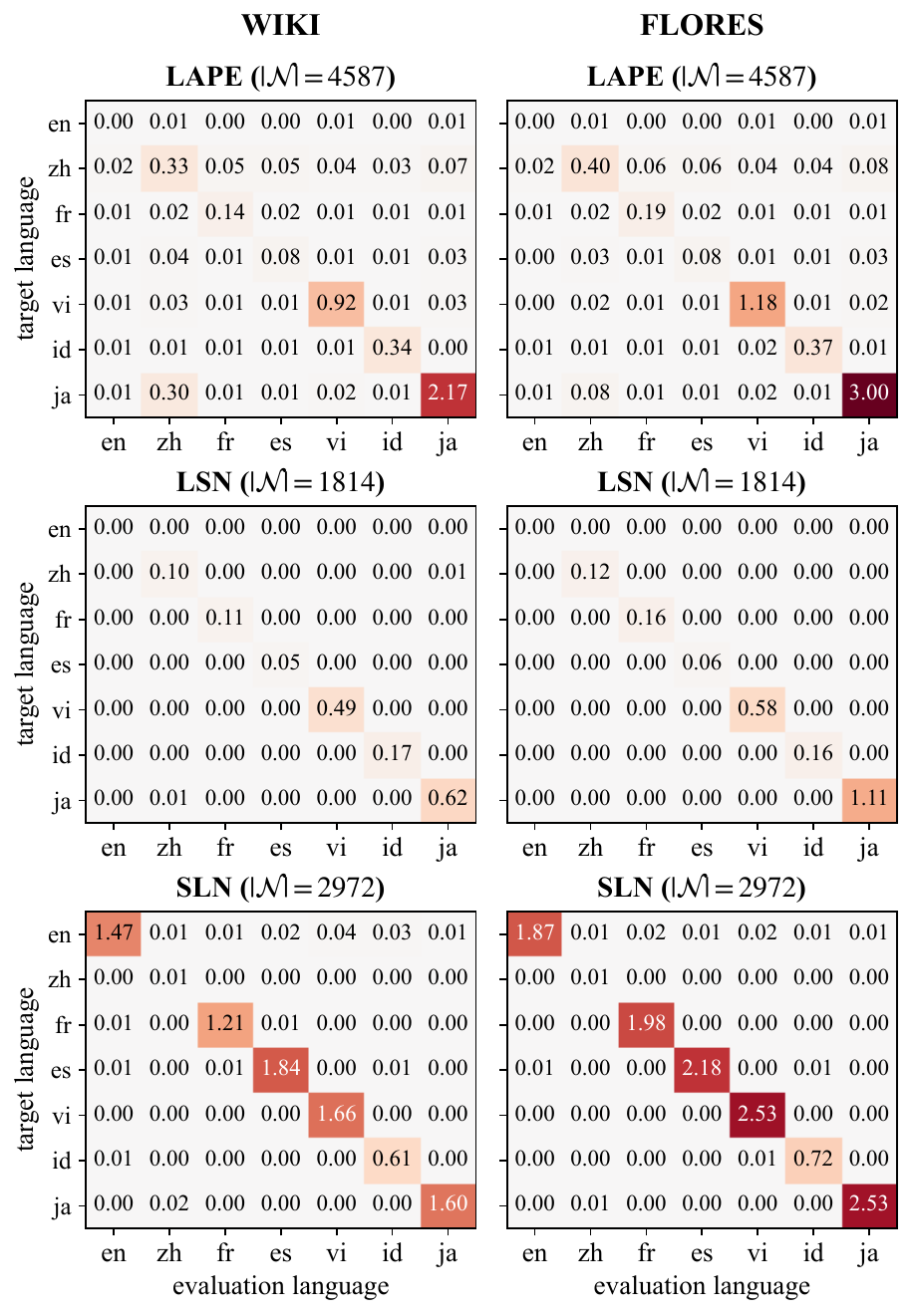}
\caption{$\Delta$NLL by language on Llama-3.1-8B for LAPE (top) and \ternarylsn\ (middle), and our SLN at $P=1\%$ (bottom) on WIKI (left) and FLORES (right). $|\mathcal{N}|$ represents the number of ablated neurons.}
\vspace{-0.1in}
\label{fig:q1_heatmap}

\end{figure}

\paragraph{Single-language neurons.} Figure~\ref{fig:q1_heatmap} compares the per-language $\Delta\mathrm{NLL}$ matrices for \lape, \ternarylsn, and SLN on Llama-3.1-8B. The three identifiers select different pool sizes ($4{,}587$ / $1{,}814$ / $2{,}972$); we therefore report on-target damage normalized by $|\mathcal{N}_{\{k\}}|$ alongside the raw $\Delta\mathrm{NLL}$. Three observations follow. 
(i) Averaged over the seven languages, SLN's on-target $\Delta\mathrm{NLL}$ per neuron is $4.9\times$ (WIKI) and $5.4\times$ (FLORES) larger than that of LAPE, and $3.1$ /  $3.2\times$ larger than \ternarylsn. 
(ii) The advantage is most pronounced for English, which both baselines fail to identify as a language-specific subset: their English-assigned subsets produce no measurable damage when ablated, whereas our SLN delivers $\Delta\mathrm{NLL}_\texttt{en}(\mathcal{N}_{\{\texttt{en}\}}) = 1.47$ (WIKI) and $1.87$ (FLORES). 
(iii) The exception is Chinese: the \texttt{zh} diagonal is weak under all three identifiers ($0.33$ / $0.10$ / $0.01$ on WIKI). Chinese activations are largely captured by the $\{$\texttt{zh}, \texttt{ja}$\}$ multi-language cluster, and the damage is recovered under the MLN setting.

\begin{figure}[t]
\centering
\includegraphics[width=\linewidth]{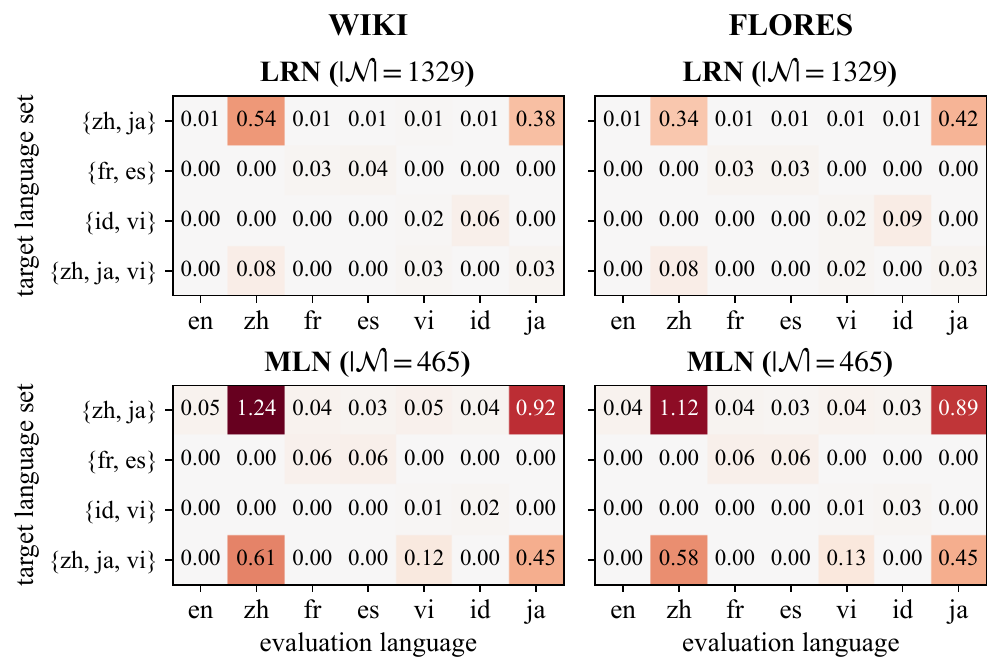}
\caption{$\Delta\mathrm{NLL}$ by cluster on Llama-3.1-8B for LRN (top) and MLN at $P{=}1\%$ (bottom) on WIKI (left) and FLORES (right). Each row corresponds to one of the top-4 most frequent MLN clusters under ablation, columns are the seven evaluation languages.}
\label{fig:q3_mln}
\end{figure}

\paragraph{Multi-language neurons.} Figure~\ref{fig:q3_mln} contrasts \ternarylrn\ with our MLN on the four most-populated cluster compositions (Table~\ref{tab:mln_neuron_counts_llama}). We make three observations.  
(i) Although our MLN uses about $2.86\times$ fewer neurons than \ternarylrn\ ($465$ vs.\ $1329$), the within-cluster damage is substantially stronger: averaged on-target $\Delta\mathrm{NLL}$ per neuron over the four clusters, our MLN delivers $6.85\times$ / $7.12\times$ (WIKI / FLORES) the damage of \ternarylrn. 
(ii) The most notable case is Chinese, which all three single-language level identifiers miss: 93.3\% of the Chinese-associated neurons identified by the proposed method are classified as MLNs (Table~\ref{tab:mln_share_p1}). 
\begin{table}[t]
    \centering
    \scriptsize
    \resizebox{\linewidth}{!}{
        \begin{tabular}{l rrrrr}
        \toprule
        Method & $\{$\texttt{zh}, \texttt{ja}$\}$ & $\{$\texttt{fr}, \texttt{es}$\}$ & $\{$\texttt{id}, \texttt{vi}$\}$ & $\{$\texttt{zh}, \texttt{ja}, \texttt{vi}$\}$\\
        \midrule
        LRN & 407 & 467 & 268 & 187 \\
        MLN & 232 & 129 &  26 &  78 \\
        \bottomrule
        \end{tabular}
    }
    \caption{Number of neurons that LRN and our MLN assign to each of the four most frequent multi-language sets on Llama-3.1-8B ($P{=}1\%$). Summed over the four sets, LRN admits $1{,}329$ neurons against our $465$.}
    \label{tab:mln_neuron_counts_llama}
\end{table}
It is the only language where a single MLN cluster accounts for the majority, with the $\{$\texttt{zh}, \texttt{ja}$\}$ cluster comprising 57.6\% of its language neurons. 
Consequently, for the $\{$\texttt{zh}, \texttt{ja}$\}$ cluster, the Chinese damage registers $\Delta\mathrm{NLL}_\texttt{zh}\left(\mathcal{N_{\{\texttt{zh,ja}\}}}\right) = 1.24 / 1.12$ on WIKI / FLORES, with within-cluster Japanese damage $0.92 / 0.89$ supporting a Chinese--Japanese coalition, reflecting their shared character-level structure. Chinese language neurons are thus not absent from the model but rather are organized as a multi-language cluster, not as a single-language set. (iii) The \{\texttt{fr}, \texttt{es}\} and \{\texttt{id}, \texttt{vi}\} clusters contribute only marginal MLN signal: for most languages, the proportion of SLNs is higher than that of MLNs. Specifically, the $\{$\texttt{fr}, \texttt{es}$\}$ cluster accounts for only 13.2\% of the \texttt{fr}-SLNs and 15.4\% of the \texttt{es}-SLNs. The $\{$\texttt{id}, \texttt{vi}$\}$ cluster also represents a small proportion compared to SLNs, comprising just 2.6\% of the \texttt{id}-SLNs and 7.5\% of the \texttt{vi}-SLNs. Therefore, these four languages are already captured at the single-language level, as their SLN diagonals reach $\Delta \mathrm{NLL}\geq 0.6$ on WIKI and $\geq 0.7$ on FLORES. Additional comparisons of SLN and MLN distributions are provided in Appendix~\ref{appendix: MLN-analysis}.

\begin{figure}[t]
\centering
\includegraphics[width=\linewidth]{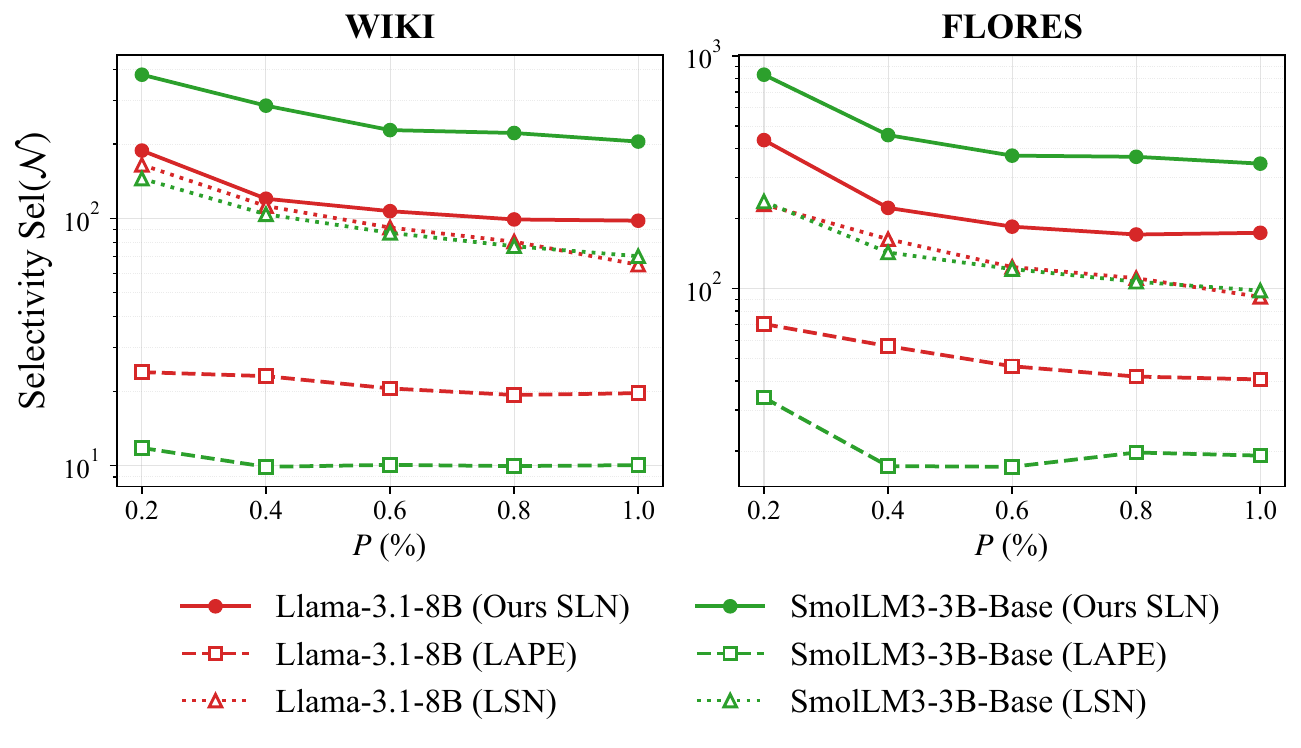}
\caption{Selectivity of SLN across the seven languages over percentile $P$. Baselines are matched to our neuron count at every $P$.}
\label{fig:q1_selectivity}
\vspace{-0.12in}
\end{figure}

\paragraph{Selectivity robustness.}
Figure~\ref{fig:q1_selectivity} reports selectivity score across the seven languages depending on $P$, for two models on both corpora. The selected neurons at percentile $P$ comprise the lowest $P$\% of clustering candidates, jointly retaining single-language ($|A_j|=1$) and multi-language ($|A_j| \in \{2, 3\}$) neurons. We match the baseline neuron budgets to ours at every $P$.

Two patterns hold consistently. First, our selectivity exceeds both LAPE and \ternarylsn\ at every $P$ in the sweep on every (model, corpus) combination: the gap over LAPE reaches an order of magnitude on SmolLM3-3B-Base ($20$--$32\times$ on WIKI, $18$--$26\times$ on FLORES) and stays at $4$--$8\times$ on Llama-3.1-8B, while the gap over \ternarylsn\ is smaller but consistent ($1.1$--$1.9\times$ on Llama-3.1-8B and $2.6$--$3.5\times$ on SmolLM3-3B-Base).  Second, selectivity decreases monotonically with $P$. Since percentile thresholding admits top ranked candidates first, smaller $P$ retains only the most concentrated subset, which is the expected behavior of a well-ordered ranking. Section~\ref{subsec:displacement} further verifies that the reported on-target damage is not a mechanical consequence of the mean-patch displacement magnitude. In an extended sweep to $P{=}5\%$, selectivity remains an order of magnitude above the budget-matched LAPE throughout (Appendix~\ref{app:psweep}).

\subsection{Effect on Downstream Tasks}
The evaluation so far measures the damage of an intervention as a change in token-level likelihood, $\Delta\mathrm{NLL}$. We now ask whether the same intervention also changes accuracy on multilingual downstream tasks. We evaluate on Belebele~\citep{bandarkar-etal-2024-belebele}, a multiple-choice reading comprehension task ($900$ questions per language), and MGSM~\citep{shi2023language}, a multilingual grade-school math benchmark ($250$ problems per language), with prompts and scoring detailed in Appendix~\ref{app:downstream_setup}. For each language neuron identification method and each target language $k$ covered by the task, we repeat the mean-patch intervention, replacing the activations of $\mathcal{N}_{\{k\}}$ with their off-target means, and report the accuracy change on every evaluation language relative to the unintervened model.

\begin{table}[t]
\centering
\small
\resizebox{\linewidth}{!}{
\begin{tabular}{l rrr rrr}
\toprule
& \multicolumn{3}{c}{\textbf{Belebele}} & \multicolumn{3}{c}{\textbf{MGSM}} \\
\cmidrule(lr){2-4} \cmidrule(lr){5-7}
\textbf{Method} & $|\mathcal{N}|$ & $k^\prime{=}k$ & $k^\prime{\neq}k$ & $|\mathcal{N}|$ & $k^\prime{=}k$ & $k^\prime{\neq}k$ \\
\midrule
\multicolumn{7}{c}{\textbf{Llama-3.1-8B}} \\
\midrule
\lape        & 4{,}587 & 2.75 & 0.07 & 3{,}088 & 6.96 & 0.88 \\
\ternarylsn  & 1{,}814 & 1.22 & $-$0.02 & 1{,}169 & 4.40 & 0.30 \\
SLN (ours)   & 2{,}972 & \textbf{4.94} & $-$0.02 & 1{,}858 & \textbf{9.44} & 0.18 \\
\midrule
\multicolumn{7}{c}{\textbf{SmolLM3-3B-Base}} \\
\midrule
\lape        & 3{,}962 & 2.46 & 0.14 & 2{,}499 & 3.60 & 0.06 \\
\ternarylsn  & 1{,}443 & 1.25 & 0.01 & \phantom{0}867 & 0.72 & $-$0.64 \\
SLN (ours)   & 2{,}254 & \textbf{5.56} & 0.15 & 1{,}331 & \textbf{7.76} & $-$0.40 \\
\bottomrule
\end{tabular}
}
\caption{Accuracy drop under the mean-patch intervention, averaged over target languages. $k^\prime{=}k$ is the on-target drop and $k^\prime{\neq}k$ the mean off-target drop. Belebele covers all seven languages, MGSM only five (\texttt{en}, \texttt{zh}, \texttt{fr}, \texttt{es}, \texttt{ja}). \textbf{Bold} denotes the largest on-target drop.
}
\vspace{-0.1in}
\label{tab:downstream}
\end{table}

Table~\ref{tab:downstream} shows that the NLL result carries over to both tasks. Notably, our SLNs induce the largest on-target drop in every (model, task) combination, while off-target accuracy remains essentially unchanged under all three identifiers. The damage is concentrated rather than indiscriminate: on Belebele, whose chance accuracy is $25.0$ and whose unintervened accuracy is $47.0$ / $45.2$ (Llama-3.1-8B / SmolLM3-3B-Base), our intervention removes $22\%$ / $27\%$ of the model's above-chance margin on the target language, yet costs no off-target language more than $2.40$ points. Our larger on-target drop is not an artifact of the number of neurons patched. Although \lape\ patches up to $1.9\times$ as many neurons as we do, our SLNs still remove more on-target accuracy per neuron than both baselines, by up to $4.0\times$ over \lape\ and $7.0\times$ over \ternarylsn. Distributional separation therefore identifies neurons whose causal effect extends beyond token-level likelihood to task behavior.

Appendix~\ref{app:downstream} reports the results for each target language, where the Chinese entry is near zero, as in the $\Delta\mathrm{NLL}$ evaluation. Our identifier classifies almost all of the Chinese language neurons as members of the $\{$\texttt{zh}, \texttt{ja}$\}$ MLN cluster, leaving the Chinese SLN pool nearly empty, so the Chinese effect surfaces under the MLN setting of Section~\ref{subsec:q1_identification}.

%% file: sections/05_analysis.tex
\section{Analysis}

\subsection{A Comparison between LAPE and SLN}
\label{sec:lape_vs_sln}

\begin{table}[t]
\centering
\resizebox{\linewidth}{!}{
\begin{tabular}{lrrrrrrr}
\toprule
\multirow{2}{*}{\textbf{Neuron Pool}} & \multirow{2}{*}{$|\mathcal{N}|$} & \multicolumn{3}{c}{\textbf{WIKI}} & \multicolumn{3}{c}{\textbf{FLORES}} \\
\cmidrule(lr){3-5} \cmidrule(lr){6-8} 
& & \textbf{$k^\prime=k$} & \textbf{$k^\prime\neq k$} & $\mathrm{Sel}(\mathcal{N})$ & \textbf{$k^\prime=k$} & \textbf{$k^\prime\neq k$} & $\mathrm{Sel}(\mathcal{N})$ \\
\midrule
\multicolumn{8}{c}{\textbf{Llama-3.1-8B}} \\
\midrule
$\text{LAPE} \setminus \text{SLN}$     & 3598         & 0.103 & 0.021 &   4.95 & 0.111 & 0.018 &   6.30 \\
$\text{LAPE} \cap \text{SLN}$          & 989          & 0.265 & 0.001 & \textbf{203.07} & 0.395 & 0.001 & \textbf{460.62} \\
$\text{SLN} \setminus \text{LAPE}$     & 1983         & 0.295 & 0.004 &  \underline{75.17} & 0.329 & 0.003 & \underline{108.83} \\
\midrule
\multicolumn{8}{c}{\textbf{SmolLM3-3B-Base}} \\
\midrule
$\text{LAPE} \setminus \text{SLN}$     & 3213         & 0.150 & 0.049 &   3.06 & 0.176 & 0.038 &   4.62 \\
$\text{LAPE} \cap \text{SLN}$          & 749          & 0.331 & 0.001 & \textbf{238.92} & 0.422 & 0.001 & \textbf{408.44} \\
$\text{SLN} \setminus \text{LAPE}$     & 1505         & 0.707 & 0.003 & \underline{204.26} & 0.878 & 0.003 & \underline{340.75} \\
\bottomrule
\end{tabular}
}
\caption{Mean selectivity across the seven languages, obtained by mean-patching the intersection and set-difference neurons of LAPE and SLN. The $k'=k$ and $k' \neq k$ columns report the average $\Delta\mathrm{NLL}_k\left(\mathcal{N}_{\{k^\prime\}}\right)$, and $\mathrm{Sel}(\mathcal{N})$ the resulting selectivity. \textbf{Bold} and \underline{underline} denote the highest and second-highest, respectively.}
\label{tab:set_comparison_lape_sln_results}
\end{table}

We first examine the following question: What are the key differences between our single-language neuron and the neurons identified by LAPE? To analyze these differences, we define and investigate the corresponding neuron pools and their sub-relations: the intersection ($\text{LAPE} \cap \text{SLN}$), and the set differences ($\text{LAPE} \setminus \text{SLN}$ and $\text{SLN} \setminus \text{LAPE}$).

\paragraph{Effect on selectivity.} We evaluate the functional importance of these distinct neuron pools by measuring their causal impact on language selectivity. As shown in Table~\ref{tab:set_comparison_lape_sln_results}, selectivity is highest for the intersection ($\text{LAPE} \cap \text{SLN}$), followed by $\text{SLN} \setminus \text{LAPE}$ and $\text{LAPE} \setminus \text{SLN}$. This implies that neurons satisfying both methods are the most effective language neurons. Notably, $\text{SLN} \setminus \text{LAPE}$ delivers a $2.86\times$ (Llama-3.1-8B) and $4.71\times$ (SmolLM3-3B-Base) higher on-target $\Delta\text{NLL}_{k}(\mathcal{N}_{\{k\}})$ than $\text{LAPE} \setminus \text{SLN}$ ($0.295$ vs.\ $0.103$; $0.707$ vs.\ $0.150$), despite using roughly half as many neurons.

\begin{figure}[t]
    \centering
    \begin{subfigure}[b]{0.49\linewidth}
        \centering
        \includegraphics[width=\linewidth]{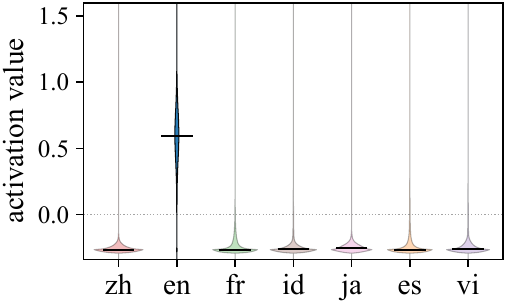}
        \caption{Only target positive}
        \label{fig:a}
    \end{subfigure}
    \hfill
    \begin{subfigure}[b]{0.49\linewidth}
        \centering
        \includegraphics[width=\linewidth]{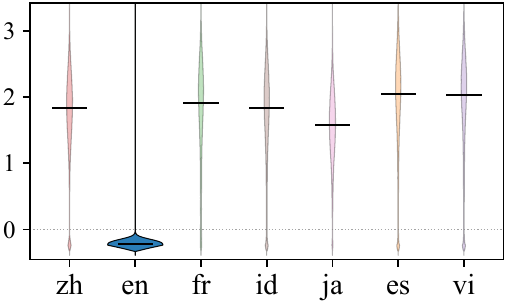}
        \caption{Only target negative}
        \label{fig:b}
    \end{subfigure}
    
    \begin{subfigure}[b]{0.49\linewidth}
        \centering
        \includegraphics[width=\linewidth]{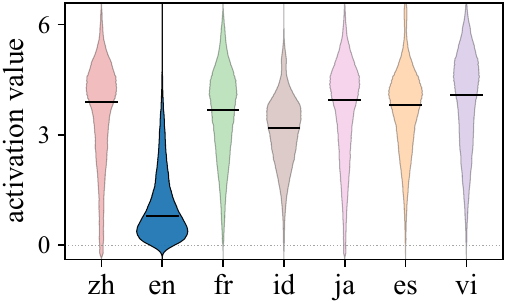}
        \caption{All positive}
        \label{fig:c}
    \end{subfigure}
    \hfill
    \begin{subfigure}[b]{0.49\linewidth}
        \centering
        \includegraphics[width=\linewidth]{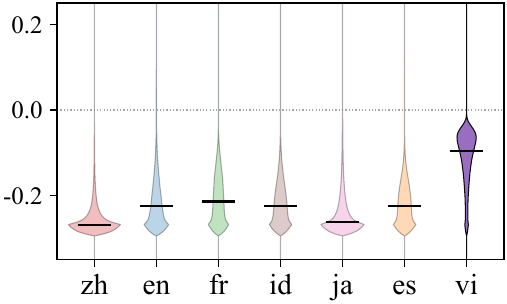}
        \caption{All negative}
        \label{fig:d}
    \end{subfigure}
    \caption{Four representative activation patterns of single language neurons across seven languages.}
    \label{fig:representative_cases}
\end{figure}

\paragraph{Distributional diversity of SLNs.}
We analyze the distributional characteristics of the SLNs identified by our method. As illustrated in Figure~\ref{fig:representative_cases}, the selected neurons can be categorized into four patterns: (\subref{fig:a}) only target positive, where only the target language has positive activations; (\subref{fig:b}) only target negative, where only the target language has negative activations; (\subref{fig:c}) all positive, where all languages have positive activations but the target language shows a distinct magnitude shift; and (\subref{fig:d}) all negative, where all activations are negative but the target language remains distributionally separated. Unlike indicator-based methods, which are limited to the target-positive pattern, our distribution-aware approach successfully uncovers this broader spectrum of language neurons. For instance, Table~\ref{tab:appendix-activation-profile} shows that 63–72\% of English SLNs have a negative target-language mean, making them entirely undetectable to positive-rate metrics despite their functional relevance.

\begin{table}[t]
\centering
\small
\begin{tabular}{llcccc}
\toprule
\multirow{2}{*}{\textbf{Direction}} & \multirow{2}{*}{\textbf{$k$}} & \multirow{2}{*}{\textbf{\lape}} & \multicolumn{3}{c}{\textbf{\ours}} \\
\cmidrule(lr){4-6}
& & & $\mathcal{N_{\{\texttt{en}\}}^{+}}$ & $\mathcal{N_{\{\texttt{en}\}}^{-}}$ & $\mathcal{N_{\{\texttt{en}\}\text{,topk}}^{-}}$ \\
\midrule
\multirow{6}{*}{\texttt{en} $\to k$}
& \texttt{zh} & 0.00 & 0.39 & \textbf{0.74} & \underline{0.72} \\
& \texttt{fr} & 0.01 & 0.00 & \textbf{0.90} & \underline{0.89} \\
& \texttt{ja} & 0.06 & 0.49 & \textbf{0.71} & \underline{0.70} \\
& \texttt{es} & 0.01 & 0.00 & \underline{0.74} & \textbf{0.78} \\
& \texttt{vi} & 0.00 & 0.00 & \underline{0.82} & \textbf{0.87} \\
& \texttt{id} & 0.00 & 0.00 & \underline{0.57} & \textbf{0.60} \\
\midrule
\multicolumn{2}{c}{Average} & 0.01 & 0.15 & \underline{0.75} & \textbf{0.76} \\
\midrule
\multirow{6}{*}{$k \to$ 	\texttt{en}}
& \texttt{zh} & \textbf{0.96} & \underline{0.95} & 0.79 & 0.66 \\
& \texttt{fr} & \underline{0.96} & \textbf{0.98} & 0.78 & 0.36 \\
& \texttt{ja} & \textbf{0.97} & \underline{0.93} & 0.65 & 0.58 \\
& \texttt{es} & \underline{0.98} & \textbf{0.99} & 0.90 & 0.54 \\
& \texttt{vi} & \textbf{0.98} & \textbf{0.98} & 0.59 & 0.21 \\
& \texttt{id} & \textbf{0.95} & \textbf{0.95} & 0.55 & 0.20 \\
\midrule
\multicolumn{2}{c}{Average} & \textbf{0.97} & \underline{0.96} & 0.71 & 0.43 \\
\bottomrule
\end{tabular}
\caption{Language accuracy of steered output. \lape\ uses its native $61$ English neurons. \ours\ splits its identified English neurons by mean-activation sign on English inputs into $\mathcal{N_{\{\texttt{en}\}}^{+}}$~(68 neurons) and $\mathcal{N_{\{\texttt{en}\}}^{-}}$~(196 neurons), and $\mathcal{N_{\{\texttt{en}\}\text{,topk}}^{-}}$ denotes the top-$68$ subset of $\mathcal{N_{\{\texttt{en}\}}^{-}}$ under the same selection ranking. \textbf{Bold} and \underline{underline} denote the highest and second-highest accuracy, respectively.}
\label{tab:steering-lid-unified}
\end{table}

\paragraph{Functional role of negative-region neurons.}\label{sec:deep_dive_into_negative}
To investigate what role these negative-region neurons serve in generation, we partition English neurons by the sign of $\mu_\text{tgt}$, the mean activation on target-language inputs, and steer generation with mean-patching in both translation directions. Source-language neurons are patched to the off-source mean and target-language neurons are patched to the target mean, thereby specifying the direction of the language switch. We evaluate 200 samples per language from each of the WIKI and FLORES datasets, yielding 400 samples per language in total. To verify whether the steered output is emitted in the intended target language, we measure language accuracy, the fraction of generations whose top-1 fastText prediction~\cite{joulin2016fasttext} matches the target, with LAPE included as a baseline. 

The results in Table~\ref{tab:steering-lid-unified} reveal a clear functional split. When the target output is English, $\mathcal{N_{\{\texttt{en}\}}^{+}}$ and \lape\ achieve near-perfect switching accuracies of $0.96$ and $0.97$, respectively, while the negative-region neurons are markedly weaker at $0.71$. This pattern reverses when the target output is a non-English language $k$: only the negative subset is effective. Specifically, $\mathcal{N_{\{\texttt{en}\}}^{-}}$ drives the switch with average language accuracy $0.75$, whereas $\mathcal{N_{\{\texttt{en}\}}^{+}}$ and \lape\ are almost ineffective with accuracies of $0.15$ and $0.01$, respectively. This effect is not merely an artifact of set size, since $\mathcal{N_{\{\texttt{en}\}\text{,topk}}^{-}}$, which is matched in size to the positive subset, retains strong performance of $0.76$, indicating that the negative-region neurons drive the switch functionally rather than through the difference in set size.

In other words, the positive-region neurons are responsible for producing English output, whereas the negative-region neurons form a functionally distinct circuit that switches away from an English context into another language. Further details are provided in Appendix~\ref{appendix:steering}. Whether such interventions can also improve task performance is a separate research problem, and \citet{mondal-etal-2025-language} report negative results for language-specific neurons facilitating cross-lingual transfer. Our scope is the faithful identification of language neurons, validated causally and on downstream tasks, rather than performance enhancement.

\subsection{Displacement-matched Control}\label{subsec:displacement}
Because mean-patching moves each neuron by the distributional separation it was selected for, one may ask whether the large on-target $\Delta\mathrm{NLL}$ merely reflects a large displacement. To test this, we apply the intervention to neurons randomly sampled from outside the DLN-selected pools and count-matched on every layer. Each is shifted from its target-language mean by the same number of standard deviations as its matched SLN ($1.6$--$2.2\sigma$; 3 seeds). This matched displacement produces only ${\sim}1\%$ of the SLN effect (mean on-target $\Delta\mathrm{NLL}$ of $0.013$ vs.\ $1.198$) and no language selectivity. Its on-target and off-target damage coincide ($0.013$ each), whereas the SLN intervention leaves off-target damage at only $0.006$. The pattern replicates on SmolLM3-3B-Base ($0.010$ vs.\ $1.395$). The reported $\Delta\mathrm{NLL}$ thus reflects which neurons are patched, not how far they are moved. Per-language results are provided in Appendix~\ref{app:displacement-control}.

\subsection{Identification across GLU Sites}\label{subsec:a_vs_z}

A GLU feed-forward block exposes two natural signals: the gate activation $a_j(x) = [\sigma(W_\mathrm{gate}^\ell x)]_n$ (the $a$-site) and the gate-up product $z_j(x) = a_j(x) \otimes u_j(x)$ (the $z$-site), where $u_j(x) = [W_\mathrm{up}^\ell x]_n$. Prior neuron-level interpretability has focused on the $a$-site for its clean above-zero criterion~\citep{tang2024language, dai-etal-2022-knowledge, geva-etal-2021-transformer}, yet the $z$-site is the signal that is causally operative. We therefore ask whether our distribution-aware identifier extends to the $z$-site. 
The two sites differ sharply in sign: pooled across neurons and the seven languages, $a$ is negative-dominant ($\mathbb{P}(a > 0) \approx 0.25$) while $z$ is sign-balanced ($\mathbb{P}(z > 0) \approx 0.50$; Table~\ref{tab:a_vs_z_sign_rates}). This matters because baselines rank neurons by the entropy of activation probabilities, which requires the activation rate to vary across languages. At the $a$-site this rate spans an approximately $3\times$ gap between the most- and least-active languages, so the entropy is informative; at the $z$-site the sign-balanced product flattens every language to $\approx 0.5$, collapsing this gap and saturating the entropy.

\begin{table}[t]
\centering
\small
\begin{tabular}{l rrr}
\toprule
& \multicolumn{3}{c}{Selection Method} \\
\cmidrule(lr){2-4}
(selection, patch) & LAPE & LSN & SLN \\
\midrule
\multicolumn{4}{c}{Llama-3.1-8B}\\
\midrule
($a$-site, $a$-site)   & 22.8 &  126.0 &  \textbf{273.7} \\
($z$-site, $a$-site)   & 4.5  &  13.3  &  \textbf{193.9} \\
($z$-site, $z$-site)   & 3.9  &  9.9  & \textbf{97.2} \\
\midrule
\multicolumn{4}{c}{SmolLM-3-3B-Base}\\
\midrule
($a$-site, $a$-site)   & 28.2 &  89.5 &  \textbf{296.5} \\
($z$-site, $a$-site)   & 7.4  &  8.8  &  \textbf{187.3} \\
($z$-site, $z$-site)   & 5.3  &  9.2  & \textbf{57.4} \\
\bottomrule
\end{tabular}
\caption{Mean selectivity across the seven languages on WIKI. Each row varies the neuron-selection site and the patching site. \textbf{Bold} denotes the highest selectivity.}
\label{tab:a_vs_z}
\end{table}

Our identifier demonstrates robustness when extending the selection from $a$ to the $z$-site, retaining most of its efficacy with a decrease of $29$--$37\%$. Consequently, at the $z$-site, our method maintains a substantial advantage, outperforming LAPE by $25$--$44\times$ and LSN by $15$--$21\times$. In contrast, the baselines lose $74$--$90\%$ of their selectivity under this transition, confirming that the performance drop is driven by the difficulty of selecting at the $z$-site, not by the change of intervention site.

In the $(z\text{-site}, z\text{-site})$ row, the cross-language mean spread at $z$-site is about $4\times$ narrower than at $a$-site (per-neuron median $0.018$ vs.\ $0.077$ on Llama-3.1-8B), so mean-patching substitutes a smaller magnitude; this drop is uniform across methods and leaves the $(z, a)$ ordering intact. That our criterion survives the move to $z$ shows distribution-aware identification applies at the causally operative signal. This locus remains underexplored under the gate-only convention and is a natural target for mechanistic analysis.

%% file: sections/06_conclusion.tex
\section{Conclusion}
In this paper, we propose a \textit{distribution-aware language neuron} selection method, which considers the full range of activation values and the pairwise relationship among language activation distributions, to better reflect the multilingual nature of mLLMs. Experimental results demonstrate that our method effectively selects neurons that influence multilingual abilities. In addition, the identified MLNs form coherent language groups whose members exhibit relatively large performance degradation when intervened on. Further analysis shows that language neurons selected from the negative activation regime play functional roles in language switching during generation. Finally, we demonstrate that our method can also be applied to other feasible locations in the model, such as the gate-up product ($z$-site) in addition to the standard gate activation. Overall, our work demonstrates that comprehensively modeling activation distributions provides an effective framework for identifying language neurons in mLLMs.

%% file: sections/98_limitations.tex
\section*{Limitations}

\paragraph{Limited language coverage.} The experiments and language neuron selection sample only seven languages (English, Chinese, French, Spanish, Vietnamese, Indonesian, and Japanese). While these cover multiple scripts and language families, they represent only a tiny fraction of the languages that modern multilingual LLMs are trained on.

\paragraph{Architectural restriction to GLU.} A key limitation of this work is its exclusive focus on autoregressive Transformer models equipped with Gated Linear Unit (GLU) feed-forward blocks with the SiLU activation function. Consequently, it remains unexplored whether this distribution-aware neuron identification generalizes effectively to models employing standard MLP architectures that lack this bifurcated signal structure.

\paragraph{Intervention penalty at the $z$-site.} When shifting identification and intervention to the $z$-site (the gate-up product), we observe an intervention-side penalty. The cross-language mean spread at this site is approximately four times smaller than at the $a$-site. As a result, the mean-patching technique substitutes a proportionally smaller magnitude, causing a uniform drop in intervention effectiveness across all evaluated methods.

%% file: sections/99.appendix.tex
\section{Implementation Details for the Baselines}
\label{app:implementation}

\citet{zhang2026does} use a balancing coefficient $\lambda$ in their score (see Baselines in Section~\ref{subsec:baselines}). It controls the trade-off between the entropy term, which favors neurons concentrated on a few languages, and the max-rate term, which favors neurons with at least one strongly-activating language.
Table~\ref{app:lambda} provides the auto-determined $\lambda_{\text{auto}}$ obtained by following their procedure.

\begin{table}[!h]
    \centering
    \small
    \begin{tabular}{lc}
    \toprule
    Model & $\lambda_\text{auto}$ \\
    \midrule
    SmolLM3-3B-Base & 0.297 \\
    Llama-3.1-8B    & 0.341 \\
    Qwen3-4B-Base   & 0.441 \\
    \bottomrule
    \end{tabular}
    \caption{$\lambda_\text{auto}$ for each model.}
    \label{app:lambda}
\end{table}

\section{Comparison under Different Interventions}
\label{app:zero-ablation}

We employ mean-patching for neuron intervention because it naturally aligns with the properties of our method, as discussed in Section~\ref{subsec:experiment_settings}. Since LAPE and \citet{zhang2026does} were originally evaluated with zero-out interventions, which align with their method designs, we additionally compare LAPE and SLN under zero-out intervention for a fair comparison. The SLN set is identical to that used in the main experiment in Figure~\ref{fig:q1_heatmap}, where $P=1\%$. For LAPE, we use the bottom $1\%$ of the entropy as the threshold, following \citet{tang2024language}.

\begin{figure}[!t]
  \centering
    \includegraphics[width=\linewidth]{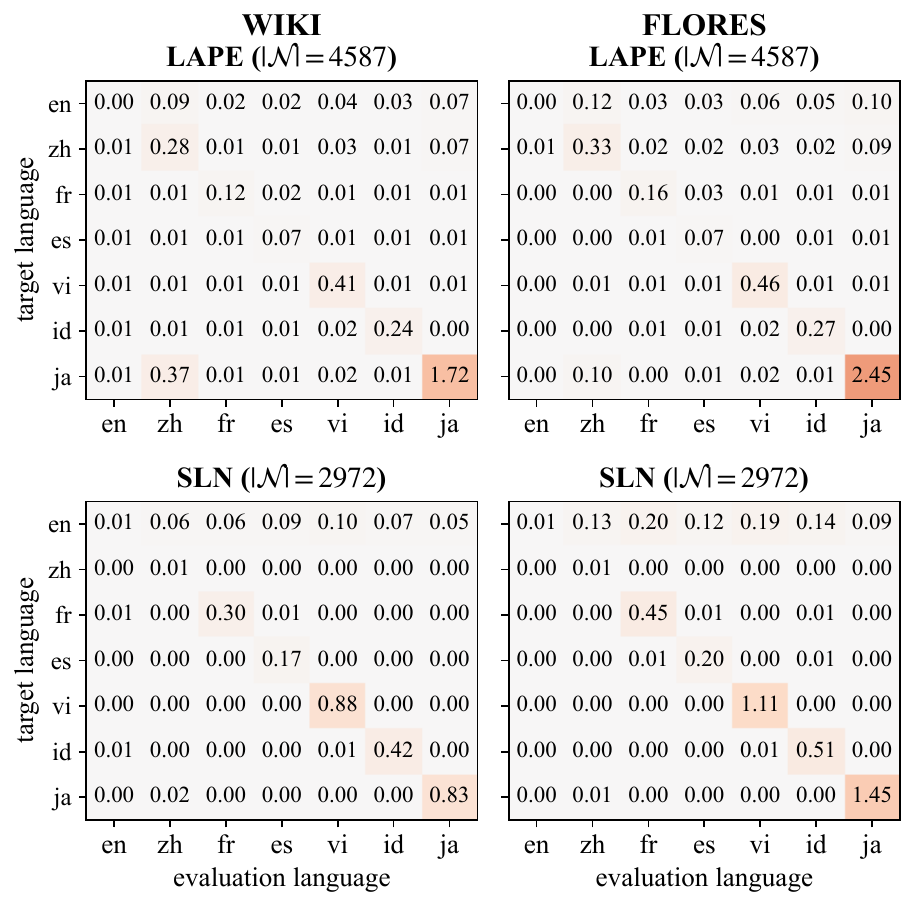}
  \caption{Per-language $\Delta\mathrm{NLL}$ under zero-ablation on Llama-3.1-8B: LAPE (top) vs SLN (bottom); WIKI (left), FLORES (right).}
  \label{fig:appendix-zeroout-llama}
\end{figure}

\begin{figure}[!t]
  \centering
    \includegraphics[width=\linewidth]{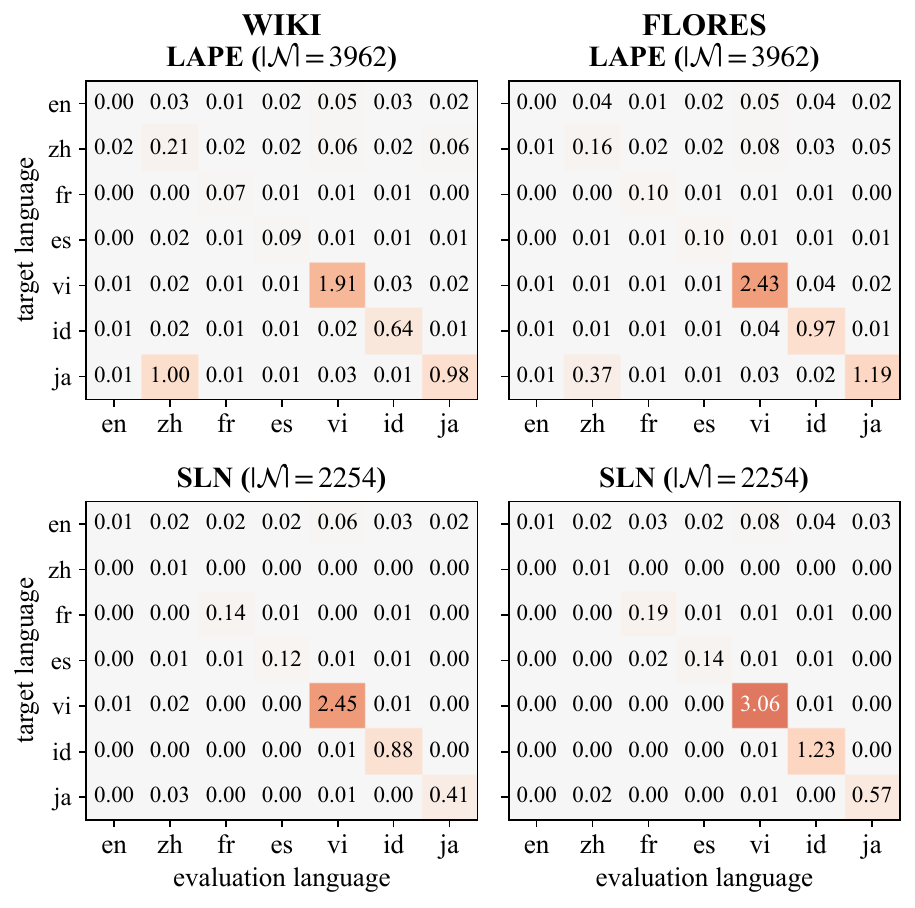}
  \caption{Per-language $\Delta\mathrm{NLL}$ under zero-ablation on SmolLM3-3B-Base: LAPE (top) vs SLN (bottom); WIKI (left), FLORES (right).}
  \label{fig:appendix-zeroout-smollm}
\end{figure}

Table~\ref{tab:appendix-zero-vs-mean-diag} shows the $\Delta$NLL results under zero-out intervention. For the zero-out intervention, our method identifies more effective language neurons than LAPE in most cases, despite selecting fewer neurons. As discussed in MLNs results in Section~\ref{subsec:q1_identification}, SLNs are less effective than LAPE under both intervention settings for languages with high coherence to other languages, such as \texttt{zh} and \texttt{ja}. However, as shown in Figures~\ref{fig:appendix-zeroout-llama}, \ref{fig:appendix-zeroout-smollm}, this is because our method separates SLNs and MLNs effectively. For example, zeroing out Chinese-specific neurons identified by LAPE also affects the model’s Japanese ability, whereas zeroing out Chinese SLNs identified by our method has little effect on Japanese.

One notable observation is the difference in $\Delta$NLL between the two interventions. For languages except \texttt{en}, both methods show relatively similar $\Delta$NLL. However, for \texttt{en}, SLN yields a much larger increase in $\Delta$NLL under mean-patching than under zero-out intervention, whereas LAPE shows little change. This suggests that English SLNs identified by our method encode language-specific information not only through positive activation but also through distributional structure in the negative activation regime. Mean-patching more directly disrupts this structure by replacing English-specific activations with non-English values, whereas zero-out may be insufficient when the relevant signal lies below zero.

\begin{table}[t]
\centering
\scriptsize
\setlength{\tabcolsep}{3.5pt}
\begin{tabular}{ll rr rr rr}
\toprule
& & \multicolumn{2}{c}{$|\mathcal{N}_k|$} & \multicolumn{2}{c}{zero $\Delta\mathrm{NLL}$} & \multicolumn{2}{c}{mean
$\Delta\mathrm{NLL}$} \\
\cmidrule(lr){3-4}\cmidrule(lr){5-6}\cmidrule(lr){7-8}
Model & Lang & LAPE & SLN & LAPE & SLN & LAPE & SLN \\
\midrule
\multirow{7}{*}{{Llama-3.1-8B}}
& \texttt{zh} & 715  & 27  & \textbf{0.280} & 0.006 & \textbf{0.332} & 0.009 \\
& \texttt{en} & 61   & 188 & 0.004 & \textbf{0.011} & 0.004 & \textbf{1.466} \\
& \texttt{fr} & 822  & 780 & 0.121 & \textbf{0.297} & 0.141 & \textbf{1.208} \\
& \texttt{id} & 1057 & 915 & 0.239 & \textbf{0.419} & 0.336 & \textbf{0.605} \\
& \texttt{ja} & 871  & 247 & \textbf{1.724} & 0.834 & \textbf{2.169} & 1.596 \\
& \texttt{es} & 619  & 616 & 0.066 & \textbf{0.165} & 0.075 & \textbf{1.843} \\
& \texttt{vi} & 442  & 199 & 0.406 & \textbf{0.881} & 0.918 & \textbf{1.662} \\
\midrule
\multirow{7}{*}{{SmolLM3-3B}}
& \texttt{zh} & 657  & 44  & \textbf{0.210} & 0.012 & \textbf{0.240} & 0.020 \\
& \texttt{en} & 45   & 152 & 0.002 & \textbf{0.009} & 0.004 & \textbf{0.506} \\
& \texttt{fr} & 539  & 475 & 0.072 & \textbf{0.136} & 0.080 & \textbf{1.096} \\
& \texttt{id} & 916  & 566 & 0.642 & \textbf{0.884} & 0.741 & \textbf{1.448} \\
& \texttt{ja} & 827  & 235 & \textbf{0.981} & 0.409 & 0.928 & \textbf{2.106} \\
& \texttt{es} & 431  & 425 & 0.085 & \textbf{0.119} & 0.094 & \textbf{1.922} \\
& \texttt{vi} & 547  & 357 & 1.907 & \textbf{2.451} & 2.036 & \textbf{2.671} \\
\midrule
\multirow{7}{*}{{Qwen3-4B}}
& \texttt{zh} & 530  & 113 & \textbf{0.150} & 0.106 & \textbf{0.146} & 0.128 \\
& \texttt{en} & 90   & 120 & 0.004 & \textbf{0.013} & 0.004 & \textbf{0.564} \\
& \texttt{fr} & 667  & 651 & 1.458 & \textbf{1.527} & \textbf{1.948} & 1.918 \\
& \texttt{id} & 529  & 690 & \textbf{1.253} & 1.091 & \textbf{1.566} & 1.191 \\
& \texttt{ja} & 850  & 663 & \textbf{3.185} & 2.968 & \textbf{3.402} & 3.235 \\
& \texttt{es} & 595  & 534 & \textbf{0.821} & 0.595 & 0.997 & \textbf{1.519} \\
& \texttt{vi} & 241  & 287 & 2.790 & \textbf{3.733} & 3.240 & \textbf{4.368} \\
\bottomrule
\end{tabular}
\caption{On-target $\Delta\mathrm{NLL}$ on WIKI after intervening on language neurons identified by LAPE and SLN under each intervention setting. \textbf{Bold} indicates the best result within each intervention setting. 
}
\label{tab:appendix-zero-vs-mean-diag}
\end{table}

\begin{table}[t]
\centering
\scriptsize
\setlength{\tabcolsep}{3.5pt}
\begin{tabular}{ll rr r rr r}
\toprule
& & \multicolumn{2}{c}{$\mu_\text{tgt}$} & $\mu_\text{off}$ & \multicolumn{2}{c}{$/\sigma$ (SLN)} & \%neg \\
\cmidrule(lr){3-4}\cmidrule(lr){5-5}\cmidrule(lr){6-7}\cmidrule(lr){8-8}
Model & Lang & LAPE & SLN & SLN & $|\mu_\text{tgt}|$ & $|\Delta\mu|$ & SLN \\
\midrule
\multirow{7}{*}{{Llama-3.1-8B}}
  & \texttt{zh} & $+$0.45 & $+$0.94 & $+$0.05 & 1.76 & 1.85 & 4\% \\
  & \texttt{en} & $+$0.44 & \underline{$+$0.05} & $+$0.69 & \underline{$0.50$} & $1.61$ & \textbf{70\%} \\
  & \texttt{fr} & $+$0.65 & $+$1.13 & $-$0.03 & 2.03 & 2.12 & 1\% \\
  & \texttt{id} & $+$0.55 & $+$1.05 & $-$0.04 & 1.94 & 2.04 & 0\% \\
  & \texttt{ja} & $+$0.52 & $+$1.07 & $-$0.04 & 2.00 & 2.17 & 2\% \\
  & \texttt{es} & $+$0.62 & $+$1.17 & $-$0.01 & 1.98 & 2.05 & 2\% \\
  & \texttt{vi} & $+$0.52 & $+$1.08 & $-$0.02 & 1.79 & 1.91 & 2\% \\
\midrule
\multirow{7}{*}{{SmolLM3-3B}}
  & \texttt{zh} & $+$0.37 & $+$0.71 & +0.01 & 1.72 & 1.98 & 5\% \\
  & \texttt{en} & $+$0.41 & \underline{$-$0.01} & $+$0.52 & \underline{0.48} & 1.57 & \textbf{72\%} \\
  & \texttt{fr} & $+$0.47 & $+$0.83 & $-$0.01 & 2.03 & 2.12 & 1\% \\
  & \texttt{id} & $+$0.29 & $+$0.58 & $-$0.02 & 1.77 & 1.89 & 2\% \\
  & \texttt{ja} & $+$0.48 & $+$0.73 & $-$0.04 & 1.81 & 2.06 & 2\% \\
  & \texttt{es} & $+$0.52 & $+$0.80 & $+$0.01 & 1.94 & 2.04 & 4\% \\
  & \texttt{vi} & $+$0.43 & $+$0.73 & $+$0.00 & 1.70 & 1.85 & 6\% \\
\midrule
\multirow{7}{*}{{Qwen3-4B}}
  & \texttt{zh} & $+$1.16 & $+$1.59 & +0.15 & 1.52 & 1.70 & 9\% \\
  & \texttt{en} & $+$1.00 & \underline{$+$0.62} & $+$0.97 & \underline{0.69} & 1.38 & \textbf{63\%} \\
  & \texttt{fr} & $+$2.88 & $+$3.28 & $-$0.04 & 2.05 & 2.11 & 2\% \\
  & \texttt{id} & $+$3.18 & $+$3.35 & $-$0.05 & 1.98 & 2.03 & 3\% \\
  & \texttt{ja} & $+$1.93 & $+$2.36 & $-$0.03 & 1.99 & 2.07 & 8\% \\
  & \texttt{es} & $+$2.88 & $+$3.23 & $-$0.04 & 2.02 & 2.09 & 2\% \\
  & \texttt{vi} & $+$3.12 & $+$2.93 & $-$0.05 & 1.90 & 1.96 & 7\% \\
\bottomrule
\end{tabular}
\caption{Per-language activation profile of the LAPE and SLN sets (wiki calibration). For each set, $\mu_\text{tgt}$ is the per-neuron mean activation under target-language inputs and $\mu_\text{off}$ is the mean over off-target inputs. The $/\sigma$ columns normalize $|\mu_\text{tgt}|$ and the language gap $|\mu_\text{off} - \mu_\text{tgt}|$ by the neuron's pooled standard deviation; \%neg is the fraction of SLN neurons with $\mu_\text{tgt}<0$. For SLN-en (underlined), the target distribution lies within $\sim\!0.5\sigma$ of zero with the majority of neurons having a negative target mean, while the language gap remains $\sim\!1.5\sigma$. This is precisely the configuration under which zero is approximately no operation on target inputs but the off-target mean is not.}
\label{tab:appendix-activation-profile}
\end{table}

Table~\ref{tab:appendix-activation-profile} locates this difference in the activation distribution itself. For English SLNs, the per-neuron mean activation under target inputs sits within $\sim\!0.5\sigma$ of zero, compared with $1.7$--$2.0\sigma$ for every other (model, language) cell. Indeed, $63$--$72\%$ of SLN neurons have a negative target-language mean. Under this configuration, setting the activation to zero leaves it within its natural target-language manifold and produces no detectable damage on target inputs; setting it to the off-target mean, which is $\sim\!1.5\sigma$ away, places it on a language-foreign value and yields the full on-target effect. This is the empirical signature of language-specificity that is encoded in distributional shift rather than in the binary event $a_j(x)>0$, and it is the structure that the indicator-based identifier family is structurally blind to.

Taken together, the zero-ablation results function as a direct empirical confirmation of the identifier--intervention coupling of Section~\ref{subsec:experiment_settings}: when our SLN set is evaluated with the LAPE-native counterfactual, the activation-rate-encoded subset of language-specificity is preserved, while the distribution-encoded subset becomes invisible. The full advantage of the distributional identifier is realized only under the matched intervention, as a consequence of two distinct kinds of language-specificity coexisting in the model. Figures~\ref{fig:appendix-zeroout-llama}--\ref{fig:appendix-zeroout-smollm} provide the full per-language $\Delta\mathrm{NLL}$ heatmaps under zero-ablation.

\section{Site-comparison: Supplementary Results}\label{app:a_vs_z}

This collects the empirical material supporting Section~\ref{subsec:a_vs_z}: per-layer sign rates that quantify the structural sign-balance of $z$ (Table~\ref{tab:a_vs_z_sign_rates}), and the full selectivity grid across both held-out corpora (Table~\ref{app:tab:a_vs_z_full}).

\paragraph{Per-layer sign rates.}
Table~\ref{tab:a_vs_z_sign_rates} lists the mean activation-positive rate $P(\cdot > 0)$ per layer, pooled across the seven calibration languages, for the post-gate signal $a$ and the gate-up product $z$. At the z-site every layer in both models satisfies $|P(z > 0) - 0.5| \le 0.002$, so the sign statistic carries no cross-layer or cross-language signal. At the a-site the rate is consistently below $0.5$ and varies widely with depth (range $[0.16, 0.48]$ on Llama-3.1-8B, $[0.18, 0.46]$ on SmolLM3-3B-Base), reflecting SiLU's negative-side asymmetry.

\begin{table*}[!h]
\centering
\small
\setlength{\tabcolsep}{7pt}
\begin{minipage}{0.5\textwidth}
\centering
\begin{tabular}{c cc cc}
\toprule
& \multicolumn{2}{c}{Llama-3.1-8B} & \multicolumn{2}{c}{SmolLM3-3B-Base} \\
\cmidrule(lr){2-3}\cmidrule(lr){4-5}
Layer & $P(a{>}0)$ & $P(z{>}0)$ & $P(a{>}0)$ & $P(z{>}0)$ \\
\midrule
0  & 0.379 & 0.500 & 0.456 & 0.500 \\
1  & 0.366 & 0.500 & 0.268 & 0.500 \\
2  & 0.343 & 0.500 & 0.247 & 0.500 \\
3  & 0.290 & 0.500 & 0.186 & 0.500 \\
4  & 0.214 & 0.500 & 0.214 & 0.501 \\
5  & 0.197 & 0.501 & 0.288 & 0.500 \\
6  & 0.192 & 0.499 & 0.268 & 0.500 \\
7  & 0.210 & 0.500 & 0.233 & 0.500 \\
8  & 0.211 & 0.499 & 0.184 & 0.501 \\
9  & 0.211 & 0.500 & 0.214 & 0.500 \\
10 & 0.211 & 0.500 & 0.217 & 0.501 \\
11 & 0.212 & 0.501 & 0.246 & 0.501 \\
12 & 0.230 & 0.499 & 0.260 & 0.499 \\
13 & 0.208 & 0.500 & 0.279 & 0.499 \\
14 & 0.196 & 0.501 & 0.272 & 0.500 \\
15 & 0.168 & 0.500 & 0.271 & 0.499 \\
16 & 0.169 & 0.500 & 0.293 & 0.500 \\
17 & 0.164 & 0.499 & 0.289 & 0.502 \\
18 & 0.176 & 0.501 & 0.260 & 0.500 \\
19 & 0.184 & 0.500 & 0.246 & 0.501 \\
\bottomrule
\end{tabular}
\end{minipage}\hfill
\begin{minipage}{0.5\textwidth}
\centering
\begin{tabular}{c cc cc}
\toprule
& \multicolumn{2}{c}{Llama-3.1-8B} & \multicolumn{2}{c}{SmolLM3-3B-Base} \\
\cmidrule(lr){2-3}\cmidrule(lr){4-5}
Layer & $P(a{>}0)$ & $P(z{>}0)$ & $P(a{>}0)$ & $P(z{>}0)$ \\
\midrule
20 & 0.192 & 0.499 & 0.284 & 0.501 \\
21 & 0.192 & 0.500 & 0.248 & 0.501 \\
22 & 0.204 & 0.500 & 0.250 & 0.499 \\
23 & 0.227 & 0.500 & 0.237 & 0.500 \\
24 & 0.249 & 0.500 & 0.222 & 0.500 \\
25 & 0.262 & 0.500 & 0.254 & 0.499 \\
26 & 0.280 & 0.500 & 0.281 & 0.500 \\
27 & 0.292 & 0.500 & 0.330 & 0.500 \\
28 & 0.308 & 0.500 & 0.333 & 0.500 \\
29 & 0.341 & 0.500 & 0.345 & 0.500 \\
30 & 0.351 & 0.499 & 0.386 & 0.500 \\
31 & 0.479 & 0.499 & 0.401 & 0.500 \\
32 & -   & -   & 0.403 & 0.500 \\
33 & -   & -   & 0.387 & 0.500 \\
34 & -   & -   & 0.389 & 0.499 \\
35 & -   & -   & 0.451 & 0.499 \\
\midrule
mean & 0.247 & 0.500 & 0.289 & 0.500 \\
min  & 0.163 & 0.499 & 0.184 & 0.499 \\
max  & 0.479 & 0.501 & 0.456 & 0.502 \\
\bottomrule
\end{tabular}
\end{minipage}
\caption{Mean activation-positive rate $P(\cdot > 0)$ per layer, pooled across the seven calibration languages, on $50{,}000$ tokens per language sampled from \texttt{WIKI-20231101}. The z-site rate stays within $0.002$ of $0.5$ across all layers in both models, while the a-site rate spans an order of magnitude in dynamic range.}
\label{tab:a_vs_z_sign_rates}
\end{table*}

\paragraph{Full selectivity grid.}
Table~\ref{app:tab:a_vs_z_full} extends Table~\ref{tab:a_vs_z} with FLORES values for all three (select-site, patch-site) configurations and all three identifiers, on both held-out corpora. Moving the selection from $a$ to $z$, indicator baselines lose $74$--$94\%$ of their selectivity across (model, corpus) combinations (LAPE $74$--$86\%$; LSN $89$--$94\%$), while our identifier retains $50$--$71\%$. At $(z,a)$ our identifier is $25$--$55\times$ stronger than LAPE and $15$--$29\times$ stronger than LSN across all (model, corpus) combinations, of the same order as the main-result a-site gap. The drop from $(z,a)$ to $(z,z)$ within each method follows the intervention-side mechanical penalty discussed in Section~\ref{subsec:a_vs_z} and applies uniformly across identifiers; it does not change the ordering established by the $(z,a)$ comparison.

\begin{table*}[!h]
\centering
\small
\setlength{\tabcolsep}{5pt}
\begin{tabular}{lll rrr}
\toprule
Model & Corpus & (selection-site, patch-site) & LAPE & LSN & SLN \\
\midrule
\multirow{6}{*}{Llama-3.1-8B}
& \multirow{3}{*}{WIKI}
 & ($a$-site, $a$-site) &  22.8 & 126.0 & \textbf{273.7} \\
& & ($z$-site, $a$-site) &   4.5 &  13.3 & \textbf{193.9} \\
& & ($z$-site, $z$-site) &   3.9 &   9.9 & \textbf{ 97.2} \\
\cmidrule{2-6}
& \multirow{3}{*}{FLORES}
 & ($a$-site, $a$-site) &  40.5 & 307.0 & \textbf{641.2} \\
& & ($z$-site, $a$-site) &   5.8 &  17.5 & \textbf{318.4} \\
& & ($z$-site, $z$-site) &   4.9 &  12.6 & \textbf{176.1} \\
\midrule
\multirow{6}{*}{SmolLM3-3B-Base}
& \multirow{3}{*}{WIKI}
 & ($a$-site, $a$-site) &  28.2 &  89.5 & \textbf{296.5} \\
& & ($z$-site, $a$-site) &   7.4 &   8.8 & \textbf{187.3} \\
& & ($z$-site, $z$-site) &   5.3 &   9.2 & \textbf{ 57.4} \\
\cmidrule{2-6}
& \multirow{3}{*}{FLORES}
 & ($a$-site, $a$-site) &  43.3 & 189.7 & \textbf{632.1} \\
& & ($z$-site, $a$-site) &   9.8 &  12.9 & \textbf{372.3} \\
& & ($z$-site, $z$-site) &   7.0 &  13.3 & \textbf{ 87.6} \\
\bottomrule
\end{tabular}
\caption{Full selectivity grid for the three identifiers across (select-site, patch-site) configurations and both held-out corpora. $(a,a)$ is the main-result protocol; $(z,a)$ varies only the selection site (the controlled comparison); $(z,z)$ varies both. Each method is evaluated at its native $P{=}1\%$ operating point. Bold entries are highest within each row.}
\label{app:tab:a_vs_z_full}
\end{table*}

\section{Further Analysis of DLN}
\subsection{Detailed Comparison with LAPE and SLN}
\begin{figure}[!h]
  \centering
    \includegraphics[width=\linewidth]{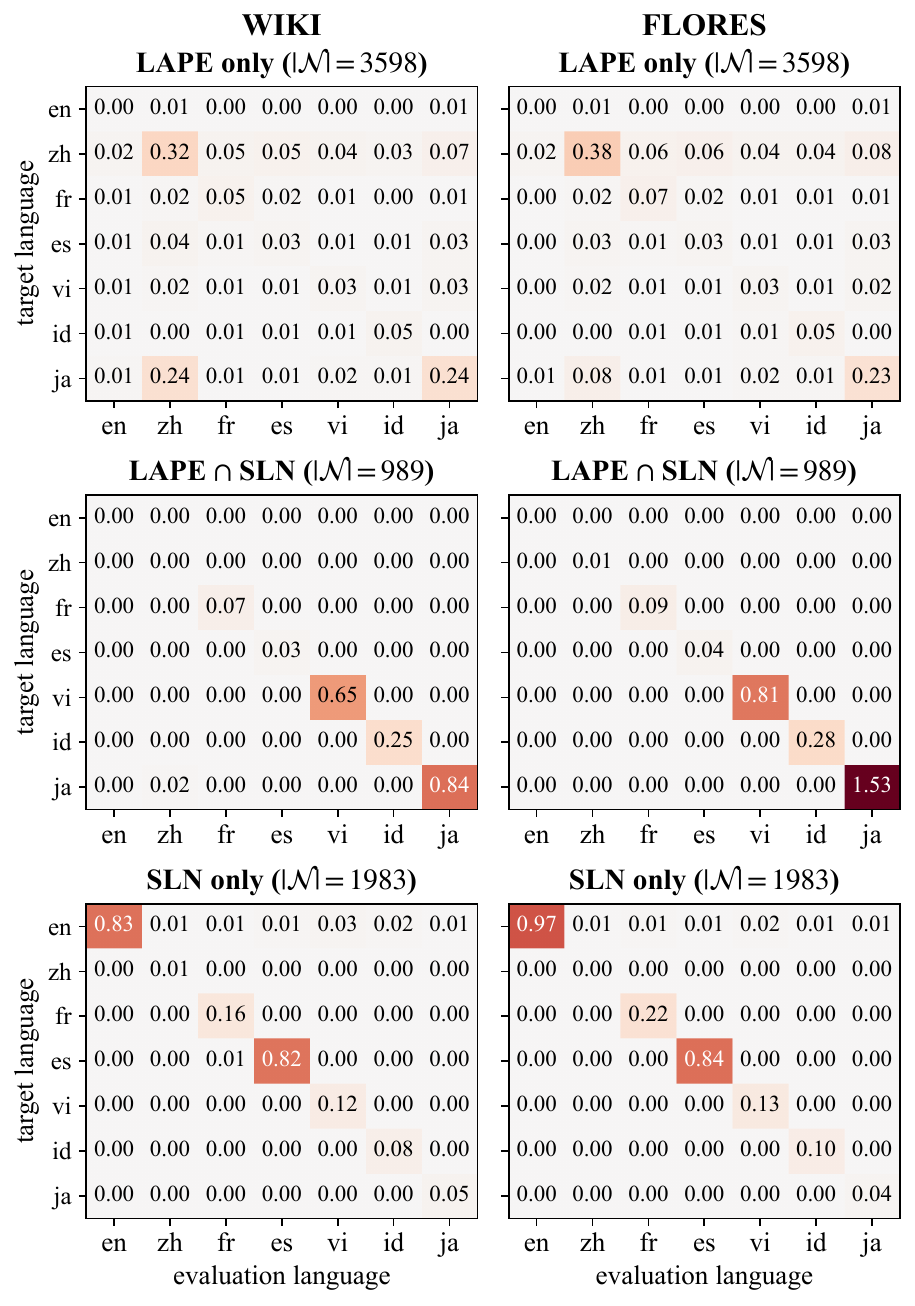}
  \caption{$\Delta$NLL by language on Llama-3.1-8B for $\text{LAPE} \setminus \text{SLN}$ (top) and $\text{LAPE} \cap \text{SLN}$ (middle), and $\text{SLN} \setminus \text{LAPE}$ single-language neuron at $P=1\%$ (bottom) on WIKI and FLORES.}
  \label{fig:appendix-heatmap-llama}
\end{figure}

\begin{figure}[!h]
  \centering
    \includegraphics[width=\linewidth]{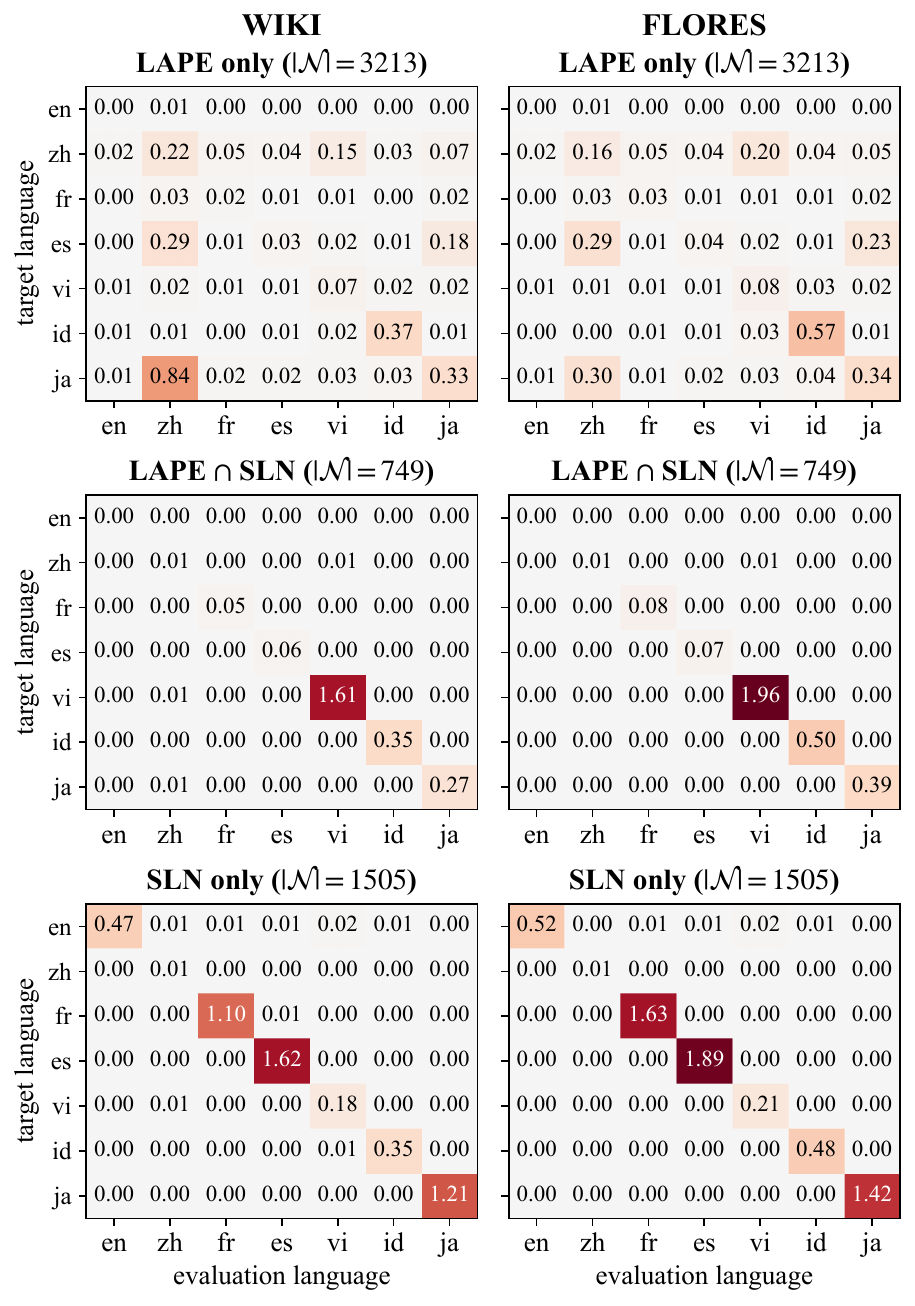}
  \caption{$\Delta$NLL by language on SmolLM3-3B-Base for $\text{LAPE} \setminus \text{SLN}$ (top) and $\text{LAPE} \cap \text{SLN}$ (middle), and $\text{SLN} \setminus \text{LAPE}$ single-language neuron at $P=1\%$ (bottom) on WIKI and FLORES.}
  \label{fig:appendix-heatmap-smollm}
\end{figure}

Figures~\ref{fig:appendix-heatmap-llama} and \ref{fig:appendix-heatmap-smollm} present heatmap results that decompose the three subsets ($\text{LAPE} \setminus \text{SLN}$, $\text{LAPE} \cap \text{SLN}$, $\text{SLN} \setminus \text{LAPE}$), which were summarized by mean selectivity in Table~\ref{tab:set_comparison_lape_sln_results}, into 7$\times$7 cells of (target language, evaluation language).

Three patterns are consistently observed across the two models. First, $\text{LAPE} \cap \text{SLN}$ (middle) exhibits the non-zero diagonal cells and near-zero off-diagonal cells, indicating that the core neurons jointly recognized by both identifiers have the most concentrated on-target effect. Second, although the neurons in $\text{SLN} \setminus \text{LAPE}$ (bottom) lie outside the LAPE pool, the magnitude of their diagonal cells exceeds that of the diagonal cells in $\text{LAPE} \setminus \text{SLN}$ (top) in nearly all languages. The gap is most pronounced in the English (\texttt{en}) cell, which is consistent with the distributional observation in Section~\ref{sec:lape_vs_sln} and Appendix~\ref{app:zero-ablation} that the target-off separation of English SLNs occurs in the negative region. Third, $\text{LAPE} \setminus \text{SLN}$ (top) contains the largest number of neurons among the three subsets, yet exhibits the weakest diagonal cells and a relatively broader off-diagonal spread, showing that the peripheral neurons of the LAPE pool, which are selected solely on the basis of activation-rate statistics, possess on average both a weaker signal and a more diffuse off-target influence.

\subsection{Impact of MLNs on Language-specific Performance} \label{appendix:mln-impact}
\begin{figure}[!h]
  \centering
    \includegraphics[width=\linewidth]{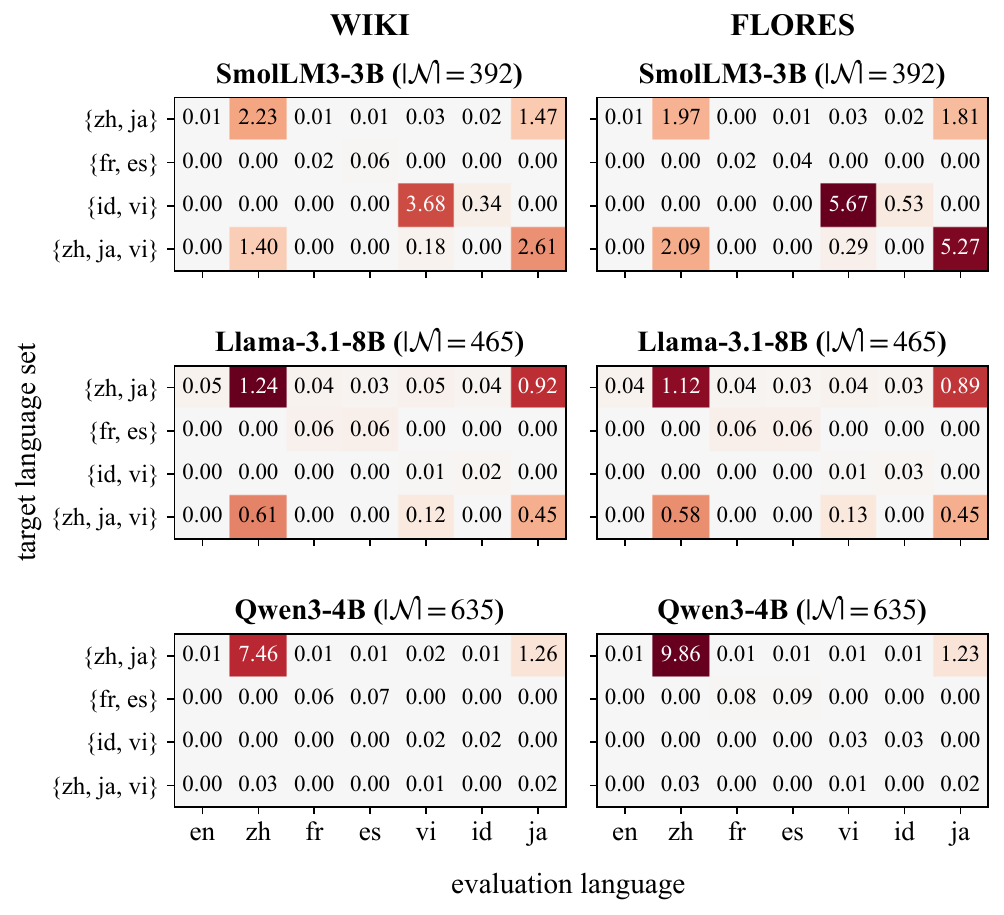}
  \caption{Comparison of $\Delta$NLL scores for the four most frequently occurring MLN clusters across Llama-3.1-8B, SmolLM3-3B-Base, and Qwen3-4B-Base. The results are evaluated on WIKI and FLORES.}
  \label{fig:appendix-mln-nll-compare}
\end{figure}

Figure~\ref{fig:appendix-mln-nll-compare} presents the $\Delta$NLL heatmaps for four of the most frequent MLN clusters across the three models. The performance of the languages included in these MLN clusters generally deteriorated in all three models.

Furthermore, a larger performance drop tends to occur when an MLN cluster accounts for a higher proportion of the neurons associated with a specific language. For instance, the $\{$\texttt{zh}, \texttt{ja}$\}$ cluster accounts for a substantial proportion of the neurons for both \texttt{zh} and \texttt{ja}, which leads to significant performance degradation. Conversely, for the $\{$\texttt{fr}, \texttt{es}$\}$ and $\{$\texttt{id}, \texttt{vi}$\}$ clusters, the performance degradation is marginal because the proportion of SLNs for each language remains higher than that of the corresponding MLNs.

However, the SmolLM3-3B-Base model presents a notable exception. Despite accounting for a relatively low proportion of the language-specific neurons, the $\{$\texttt{id}, \texttt{vi}$\}$ MLN cluster still caused a significant performance degradation. This phenomenon can be attributed to the fact that these languages are not among the model's officially supported languages (English, French, Spanish, German, Italian, and Portuguese), resulting in relatively unstable signals compared to the target languages. In other words, for unsupported languages, crucial information may heavily rely on a small subset of neurons, suggesting that MLNs likely encompass the neurons essential for processing these languages. Consequently, this demonstrates that the nature and importance of the information encoded within MLNs can vary depending on the model and its training data. Therefore, to fully comprehend language-specific mechanisms, an in-depth analysis of MLNs, alongside SLNs, is essential.

We scaled up the above experiment to a larger model, revealing a consistent trend. 
Figure~\ref{fig:appendix-mln-nll-compare-llama70b} presents the $\Delta$NLL heatmap for the four most frequently observed MLN clusters in Llama-3.1-70B. For the $\{$\texttt{zh}, \texttt{ja}$\}$ cluster, which accounts for a high proportion of MLNs in its respective languages, a substantial performance drop is observed. Conversely, the $\{$\texttt{fr}, \texttt{es}$\}$ cluster, characterized by a lower MLN and higher SLN proportion, exhibits a relatively minor degradation. A similar pattern is evident in the $\{$\texttt{en}, \texttt{fr}, \texttt{es}$\}$ cluster, where the MLN share within each language is too minimal to cause performance drop. These findings demonstrate that our proposed methodology and the observed MLN dynamics are not constrained by model scale, highlighting their broad applicability.

\subsection{Scaling to Llama-3.1-70B}
\begin{figure}[!h]
  \centering
    \includegraphics[width=\linewidth]{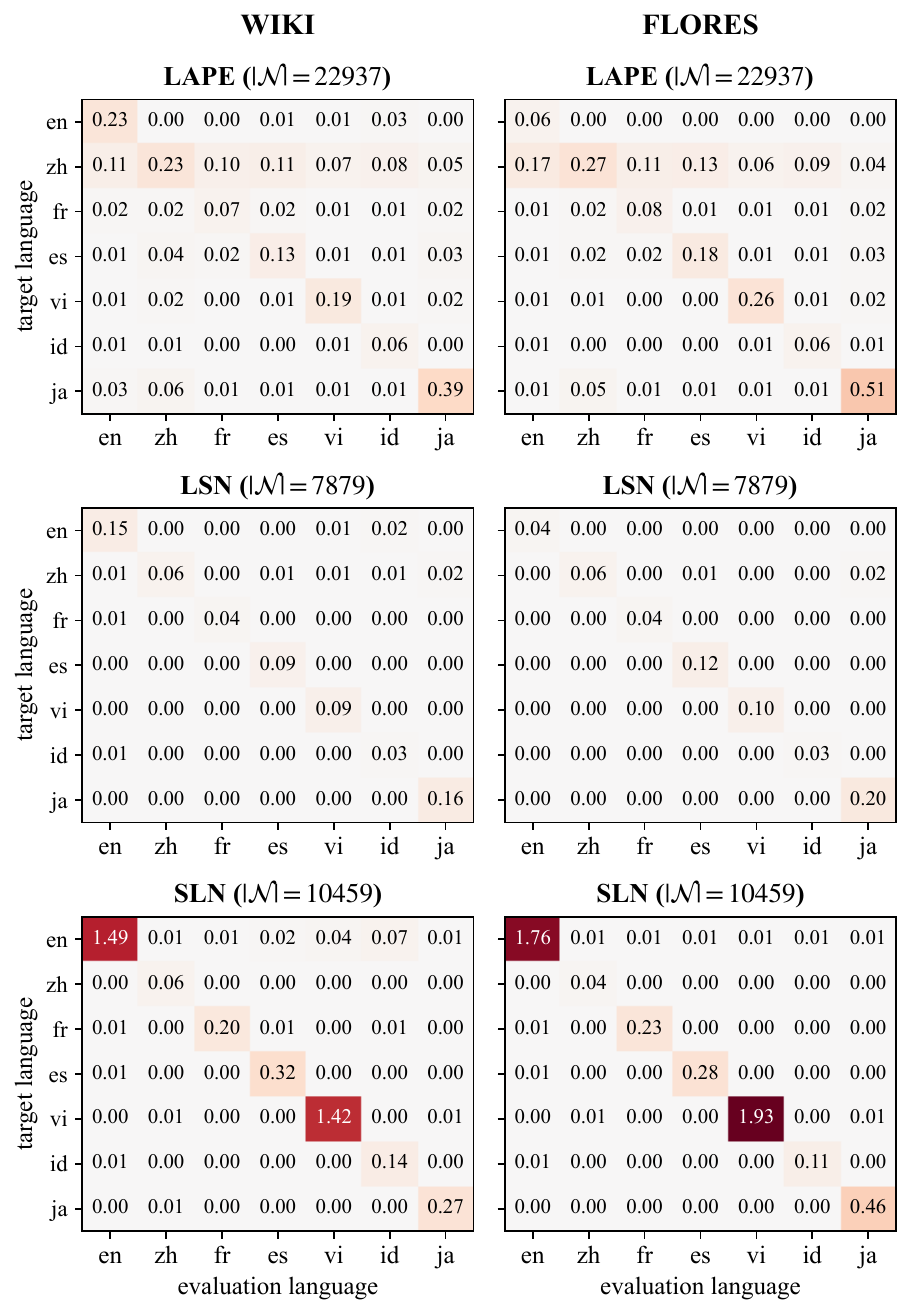}
  \caption{$\Delta$NLL by language on Llama-3.1-70B for LAPE, LSN, and SLN at $P=1\%$ on WIKI and FLORES.}
  \label{fig:appendix-heatmap-llama70b}
  \vspace{-0.1in}
\end{figure}

\begin{figure}[!h]
  \centering
    \includegraphics[width=\linewidth]{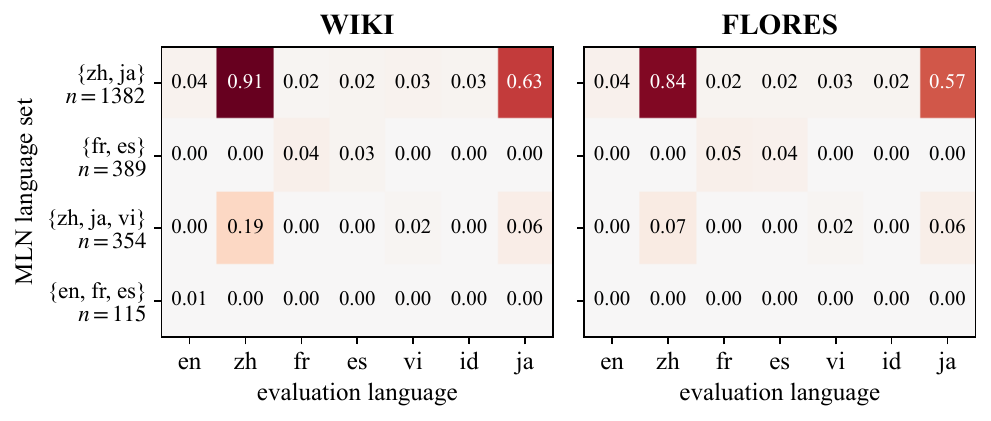}
  \caption{Comparison of $\Delta$NLL scores for the four most frequently occurring MLN clusters of Llama-3.1-70B. The results are evaluated on WIKI and FLORES.}
  \label{fig:appendix-mln-nll-compare-llama70b}
\end{figure}

Figure~\ref{fig:appendix-heatmap-llama70b} reports the SLN comparison on Llama-3.1-70B at $P{=}1\%$. The findings from the smaller models carry over. Our SLN concentrates far larger on-target damage than LAPE and LSN while using less than half the neurons of LAPE, and its off-target damage stays near zero. English, where both baselines remain weak ($0.23$ and $0.15$ on WIKI), is again recovered only by SLN ($1.49$ on WIKI, $1.76$ on FLORES). The one weak diagonal, Chinese, mirrors the 8B models: its signal is organized in the $\{$\texttt{zh}, \texttt{ja}$\}$ MLN cluster (Figure~\ref{fig:appendix-mln-nll-compare-llama70b}).

\subsection{Generalization to Aya-23-8B}
\label{appendix:aya}

We run the full pipeline on Aya-23-8B~\citep{aryabumi2024aya}, an instruction-tuned model trained explicitly for multilinguality, under the identical protocol ($P{=}1\%$; LAPE $4{,}587$ / LSN $1{,}618$ / SLN $3{,}800$ neurons). Table~\ref{tab:aya} reports per-language on-target $\Delta\mathrm{NLL}$. All findings replicate: our SLN produces the largest on-target damage on all seven languages and both corpora, while its mean off-target damage remains the lowest ($0.006$ vs.\ LAPE's $0.012$ on WIKI), yielding mean selectivity $254$ against $133$ (LSN) and $24$ (LAPE) on WIKI, and $691$ against $350$ and $38$ on FLORES.

The multi-language structure is likewise stable. The $\{$\texttt{zh}, \texttt{ja}$\}$ coalition is again the largest MLN cluster ($355$ neurons), followed by $\{$\texttt{fr}, \texttt{es}$\}$ and $\{$\texttt{id}, \texttt{vi}$\}$, and Chinese is the one language whose SLN diagonal is weak ($0.25$); the union pool recovers it ($2.36$ on WIKI), with the extra damage concentrated on its cluster-mate Japanese, exactly as the framework predicts. Since no base checkpoint of Aya-23 is released, these results show that the method transfers to post-trained models.

\begin{table}[t]
    \centering
    \small
    \setlength{\tabcolsep}{4pt}
    \begin{tabular}{l rrr rrr}
    \toprule
    & \multicolumn{3}{c}{WIKI} & \multicolumn{3}{c}{FLORES} \\
    \cmidrule(lr){2-4} \cmidrule(lr){5-7}
    lang & LAPE & LSN & SLN & LAPE & LSN & SLN \\
    \midrule
    \texttt{en} & 0.004 & 0.004 & \textbf{1.835} & 0.006 & 0.005 & \textbf{2.569} \\
    \texttt{zh} & 0.166 & 0.057 & \textbf{0.248} & 0.129 & 0.055 & \textbf{0.207} \\
    \texttt{fr} & 0.191 & 0.127 & \textbf{1.986} & 0.297 & 0.185 & \textbf{3.204} \\
    \texttt{es} & 0.121 & 0.092 & \textbf{1.736} & 0.143 & 0.114 & \textbf{2.169} \\
    \texttt{vi} & 0.623 & 0.385 & \textbf{1.213} & 0.802 & 0.451 & \textbf{1.583} \\
    \texttt{id} & 0.481 & 0.240 & \textbf{1.472} & 0.727 & 0.331 & \textbf{2.356} \\
    \texttt{ja} & 0.351 & 0.173 & \textbf{2.528} & 0.382 & 0.201 & \textbf{3.394} \\
    \midrule
    mean & 0.277 & 0.154 & \textbf{1.574} & 0.355 & 0.192 & \textbf{2.212} \\
    \bottomrule
    \end{tabular}
    \caption{Per-language on-target $\Delta\mathrm{NLL}$ on Aya-23-8B at $P{=}1\%$. \textbf{Bold} denotes the largest damage per (language, corpus).}
    \label{tab:aya}
\end{table}

\subsection{Generation Validation via Cross-lingual Steering}
\label{appendix:steering}

To analyze the causal impact of the identified language-specific neurons on text generation, we employ activation steering during inference. 
We apply this steering to the neurons extracted by each respective methodology, utilizing a paired intervention technique that simultaneously manipulates two distinct sets of neurons: source SLNs and target SLNs. For the source SLNs, we suppress their language-specific signals by applying mean-patching (Section~\ref{subsec:experiment_settings}), which replaces their activations with a language-neutral baseline. However, to ensure a fair comparison with LAPE, zero-patching, which deactivates the neurons by setting their values to zero, is applied to LAPE instead of mean-patching. Concurrently, the target language neurons are activated by injecting their pre-computed average activation values specific to the target language. When applying this technique to MLNs, any overlap between source and target neurons is resolved by prioritizing the target neuron activation. Ultimately, this dual intervention effectively spoofs the model's internal state, mimicking the natural activation patterns in the target language.

\begin{table}[t]
\centering
\small
\begin{tabular}{l|cc|cc|cc}
\toprule
 & \multicolumn{2}{c}{LAPE} & \multicolumn{2}{c}{SLN} & \multicolumn{2}{c}{SLN+MLN} \\
lang & ACC & $p_{tgt}$ & ACC & $p_{tgt}$ & ACC & $p_{tgt}$ \\
\midrule
\multicolumn{7}{c}{\textbf{SmolLM3-3B}} \\
\midrule
\texttt{zh} & \textbf{1.000} & \textbf{0.963} & \textbf{1.000} & 0.935 & \textbf{1.000} & 0.951 \\
\texttt{ja} & \textbf{1.000} & \textbf{0.955} & 0.970 & 0.895 & 0.980 & 0.867 \\
\texttt{vi} & \textbf{1.000} & \textbf{0.965} & \textbf{1.000} & 0.962 & \textbf{1.000} & 0.950 \\
\texttt{id} & \textbf{1.000} & 0.946 & \textbf{1.000} & 0.955 & \textbf{1.000} & \textbf{0.958} \\
\texttt{es} & \textbf{1.000} & \textbf{0.948} & \textbf{1.000} & 0.922 & \textbf{1.000} & 0.934 \\
\texttt{fr} & \textbf{1.000} & \textbf{0.964} & \textbf{1.000} & 0.953 & \textbf{1.000} & 0.956 \\
\cmidrule(lr){1-7}
avg & \textbf{1.000} & \textbf{0.957} & 0.995 & 0.937 & 0.997 & 0.936 \\
\midrule
\multicolumn{7}{c}{\textbf{Llama-3.1-8B}} \\
\midrule
\texttt{zh} & \textbf{1.000} & \textbf{0.958} & 0.990 & 0.920 & \textbf{1.000} & 0.874 \\
\texttt{ja} & \textbf{1.000} & \textbf{0.925} & \textbf{1.000} & 0.865 & \textbf{1.000} & 0.899 \\
\texttt{vi} & \textbf{1.000} & \textbf{0.970} & \textbf{1.000} & 0.968 & \textbf{1.000} & 0.966 \\
\texttt{id} & \textbf{1.000} & 0.964 & \textbf{1.000} & 0.964 & \textbf{1.000} & \textbf{0.965} \\
\texttt{es} & \textbf{1.000} & 0.961 & \textbf{1.000} & 0.962 & \textbf{1.000} & \textbf{0.964} \\
\texttt{fr} & 0.990 & 0.948 & \textbf{1.000} & 0.962 & \textbf{1.000} & \textbf{0.966} \\
\cmidrule(lr){1-7}
avg & 0.998 & \textbf{0.954} & 0.998 & 0.940 & \textbf{1.000} & 0.939 \\
\midrule
\multicolumn{7}{c}{\textbf{Qwen3-4B}} \\
\midrule
\texttt{zh} & 0.968 & 0.869 & 0.990 & 0.918 & \textbf{1.000} & \textbf{0.944} \\
\texttt{ja} & 0.959 & 0.886 & 0.934 & 0.800 & \textbf{1.000} & \textbf{0.937} \\
\texttt{vi} & 0.940 & 0.870 & \textbf{1.000} & 0.955 & \textbf{1.000} & \textbf{0.968} \\
\texttt{id} & \textbf{1.000} & 0.942 & \textbf{1.000} & \textbf{0.974} & \textbf{1.000} & 0.952 \\
\texttt{es} & 0.970 & 0.941 & \textbf{1.000} & \textbf{0.965} & 0.990 & 0.961 \\
\texttt{fr} & 0.990 & 0.953 & \textbf{1.000} & 0.963 & \textbf{1.000} & \textbf{0.972} \\
\cmidrule(lr){1-7}
avg & 0.971 & 0.910 & 0.987 & 0.929 & \textbf{0.998} & \textbf{0.956} \\
\bottomrule
\end{tabular}
\caption{Performance comparison of LAPE, SLN, and SLN+MLN on the X-to-English (X2en) transfer task. The table reports accuracy and target language probability ($p_{tgt}$) across three models. Zero-patching is applied to LAPE, whereas mean-patching is utilized for our proposed SLN and SLN+MLN interventions. \textbf{Bold} indicates the highest accuracy.}
\label{tab:x2en-lape-vs-ours}
\end{table}
For the experimental setup, we extract 200 samples per language from the FLORES dataset. To ensure rigorous language verification, we utilize the fastText language identification model~\cite{joulin2016fasttext}, selecting only those samples where the predicted probability for the target language is 0.98 or higher.

Table~\ref{tab:x2en-lape-vs-ours} presents the results of transferring text from a non-English source language $X$ to English as the target language. The performance discrepancies are marginal, with an average accuracy difference of 0.027 and a maximum per-language difference of 0.066. This robust performance stems from the models being predominantly pre-trained on English. The models frequently default to generating English responses even when provided with non-English inputs, naturally leading to consistently high baseline scores.

To evaluate cross-lingual transfer capabilities toward a non-primary language, we conduct experiments transferring from a source language $X$ to Chinese (\texttt{zh}) as the target language. As demonstrated in Table~\ref{tab:x2zh-lape-vs-ours}, \ours{} consistently outperforms LAPE across all scenarios. Notably, a substantial performance gap is observed between SLNs and MLNs. As analyzed in Appendix~\ref{appendix: MLN-analysis} and \ref{appendix:mln-impact}, this gap arises because Chinese-specific textual features are predominantly processed within MLNs.

\begin{table}[t]
\centering
\small
\begin{tabular}{l|cc|cc|cc}
\toprule
 & \multicolumn{2}{c}{LAPE} & \multicolumn{2}{c}{SLN} & \multicolumn{2}{c}{SLN+MLN} \\
lang & ACC & $p_{tgt}$ & ACC & $p_{tgt}$ & ACC & $p_{tgt}$ \\
\midrule
\multicolumn{7}{c}{\textbf{SmolLM3-3B}} \\
\midrule
\texttt{en} & 0.000 & 0.004 & 0.000 & 0.001 & \textbf{0.845} & \textbf{0.802} \\
\texttt{ja} & 0.236 & 0.211 & \textbf{0.990} & \textbf{0.941} & \textbf{0.990} & 0.922 \\
\texttt{vi} & 0.970 & 0.941 & 0.178 & 0.136 & \textbf{1.000} & \textbf{0.968} \\
\texttt{id} & 0.000 & 0.000 & 0.000 & 0.000 & \textbf{1.000} & \textbf{0.985} \\
\texttt{es} & \textbf{0.969} & \textbf{0.937} & 0.000 & 0.000 & 0.929 & 0.904 \\
\texttt{fr} & 0.863 & 0.779 & 0.000 & 0.000 & \textbf{0.917} & \textbf{0.861} \\
\cmidrule(lr){1-7}
avg & 0.506 & 0.479 & 0.195 & 0.180 & \textbf{0.947} & \textbf{0.907} \\
\midrule
\multicolumn{7}{c}{\textbf{Llama-3.1-8B}} \\
\midrule
\texttt{en} & 0.000 & 0.000 & 0.000 & 0.000 & \textbf{0.891} & \textbf{0.791} \\
\texttt{ja} & 0.768 & 0.694 & 0.811 & 0.755 & \textbf{0.955} & \textbf{0.795} \\
\texttt{vi} & \textbf{0.910} & \textbf{0.764} & 0.011 & 0.014 & 0.880 & 0.693 \\
\texttt{id} & \textbf{0.875} & \textbf{0.806} & 0.000 & 0.000 & 0.375 & 0.417 \\
\texttt{es} & 0.705 & 0.599 & 0.000 & 0.000 & \textbf{0.863} & \textbf{0.801} \\
\texttt{fr} & \textbf{0.652} & 0.535 & 0.000 & 0.000 & 0.606 & \textbf{0.582} \\
\cmidrule(lr){1-7}
avg & 0.652 & 0.567 & 0.137 & 0.128 & \textbf{0.762} & \textbf{0.680} \\
\midrule
\multicolumn{7}{c}{\textbf{Qwen3-4B}} \\
\midrule
\texttt{en} & 0.900 & 0.845 & \textbf{1.000} & 0.980 & \textbf{1.000} & \textbf{0.987} \\
\texttt{ja} & 0.979 & \textbf{0.906} & \textbf{1.000} & 0.850 & 0.989 & 0.848 \\
\texttt{vi} & \textbf{1.000} & 0.940 & 0.920 & 0.910 & \textbf{1.000} & \textbf{0.993} \\
\texttt{id} & \textbf{1.000} & \textbf{0.995} & 0.500 & 0.521 & \textbf{1.000} & 0.987 \\
\texttt{es} & 0.979 & 0.935 & 0.660 & 0.653 & \textbf{1.000} & \textbf{0.977} \\
\texttt{fr} & 0.907 & 0.868 & 0.320 & 0.319 & \textbf{1.000} & \textbf{0.979} \\
\cmidrule(lr){1-7}
avg & 0.961 & 0.915 & 0.733 & 0.705 & \textbf{0.998} & \textbf{0.962} \\
\bottomrule
\end{tabular}
\caption{Performance comparison of LAPE, SLN, and SLN+MLN on the X-to-Chinese (X2zh) transfer task. The table reports accuracy and target language probability ($p_{tgt}$) across three models. Zero-patching is applied to LAPE, whereas mean-patching is utilized for our proposed SLN and SLN+MLN interventions. \textbf{Bold} indicates the highest accuracy.}
\label{tab:x2zh-lape-vs-ours}
\end{table}

Table~\ref{tab:en2x-lape-vs-ours} demonstrates the performance when transferring from English as the source language to all other target languages. In stark contrast to the X-to-English transfer task, the performance of LAPE approaches zero. Conversely, \ours{} maintains solid performance, with accuracy ranging from 0.599 to 0.952 depending on the architecture. This phenomenon is primarily driven by the negative regions analyzed in Section~\ref{sec:deep_dive_into_negative}. However, our method exhibits lower performance for Vietnamese (\texttt{vi}) and Indonesian (\texttt{id}) within the SmolLM3-3B architecture, which stems from unstable internal signals provided by the model, as discussed in Appendix~\ref{appendix:mln-impact}. This explanation is supported by the results of Llama-3.1-8B, which underwent extensive multilingual pre-training and consistently delivers stable, high-tier performance.

\begin{table}[t]
\centering
\small
\begin{tabular}{l|cc|cc|cc}
\toprule
 & \multicolumn{2}{c}{LAPE} & \multicolumn{2}{c}{SLN} & \multicolumn{2}{c}{SLN+MLN} \\
 
lang & ACC & $p_{tgt}$ & ACC & $p_{tgt}$ & ACC & $p_{tgt}$ \\
\midrule
\multicolumn{7}{c}{\textbf{SmolLM3-3B}} \\
\midrule
\texttt{zh} & 0.000 & 0.004 & 0.000 & 0.001 & \textbf{0.845} & \textbf{0.802} \\
\texttt{ja} & 0.000 & 0.001 & 0.000 & 0.001 & \textbf{0.854} & \textbf{0.814} \\
\texttt{vi} & 0.000 & 0.000 & 0.010 & 0.011 & \textbf{0.414} & \textbf{0.406} \\
\texttt{id} & 0.000 & 0.000 & 0.051 & 0.038 & \textbf{0.131} & \textbf{0.096} \\
\texttt{es} & 0.000 & 0.001 & \textbf{0.717} & \textbf{0.671} & 0.653 & 0.615 \\
\texttt{fr} & 0.000 & 0.001 & \textbf{0.788} & \textbf{0.765} & 0.697 & 0.674 \\
\cmidrule(lr){1-7}
avg & 0.000 & 0.001 & 0.261 & 0.248 & \textbf{0.599} & \textbf{0.568} \\
\midrule
\multicolumn{7}{c}{\textbf{Llama-3.1-8B}} \\
\midrule
\texttt{zh} & 0.000 & 0.000 & 0.000 & 0.000 & \textbf{0.891} & \textbf{0.791} \\
\texttt{ja} & 0.020 & 0.023 & 0.060 & 0.061 & \textbf{0.956} & \textbf{0.860} \\
\texttt{vi} & 0.000 & 0.000 & 0.927 & 0.889 & \textbf{0.947} & \textbf{0.892} \\
\texttt{id} & 0.000 & 0.000 & \textbf{0.735} & \textbf{0.610} & 0.714 & 0.543 \\
\texttt{es} & 0.000 & 0.000 & \textbf{0.840} & \textbf{0.803} & 0.780 & 0.734 \\
\texttt{fr} & 0.010 & 0.010 & \textbf{0.970} & \textbf{0.941} & 0.900 & 0.871 \\
\cmidrule(lr){1-7}
avg & 0.005 & 0.006 & 0.589 & 0.551 & \textbf{0.865} & \textbf{0.782} \\
\midrule
\multicolumn{7}{c}{\textbf{Qwen3-4B}} \\
\midrule
\texttt{zh} & 0.900 & 0.845 & \textbf{1.000} & 0.980 & \textbf{1.000} & \textbf{0.987} \\
\texttt{ja} & 0.990 & 0.970 & \textbf{1.000} & \textbf{0.999} & \textbf{1.000} & 0.996 \\
\texttt{vi} & 0.040 & 0.046 & 0.660 & 0.657 & \textbf{0.990} & \textbf{0.990} \\
\texttt{id} & 0.010 & 0.004 & 0.450 & 0.364 & \textbf{0.800} & \textbf{0.669} \\
\texttt{es} & 0.010 & 0.015 & 0.420 & 0.390 & \textbf{0.930} & \textbf{0.866} \\
\texttt{fr} & 0.250 & 0.222 & 0.520 & 0.498 & \textbf{0.990} & \textbf{0.960} \\
\cmidrule(lr){1-7}
avg & 0.367 & 0.350 & 0.675 & 0.648 & \textbf{0.952} & \textbf{0.911} \\
\bottomrule
\end{tabular}
\caption{Performance comparison of LAPE, SLN, and SLN+MLN on the English-to-X (en2X) transfer task. The table reports accuracy and target language probability ($p_{tgt}$) across three models. Zero-patching is applied to LAPE, whereas mean-patching is utilized for our proposed SLN and SLN+MLN interventions. \textbf{Bold} indicates the highest accuracy.}
\label{tab:en2x-lape-vs-ours}
\end{table}

\subsection{Extended Percentile Sweep}
\label{app:psweep}

\begin{table}[!h]
    \centering
    \small
    \setlength{\tabcolsep}{5pt}
    \begin{tabular}{r r rr r r}
    \toprule
    & & \multicolumn{3}{c}{ours} & LAPE \\
    \cmidrule(lr){3-5} \cmidrule(lr){6-6}
    $P$ (\%) & $|\mathcal{N}|$ & on-tgt & off-tgt & $\mathrm{Sel}$ & $\mathrm{Sel}$ \\
    \midrule
    1.5 & 3,885 & 0.403 & 0.003 & 145.8 & 10.3 \\
    3.0 & 7,336 & 0.322 & 0.004 & 88.7 & 9.8 \\
    5.0 & 12,444 & 0.294 & 0.004 & 69.4 & 8.1 \\
    \bottomrule
    \end{tabular}
    \caption{Extended percentile sweep of the DLN pool on SmolLM3-3B-Base (WIKI). LAPE is matched to our neuron budget $|\mathcal{N}|$ at every $P$.}
    \label{tab:psweep}
\end{table}

Figure~\ref{fig:q1_selectivity} sweeps the percentile threshold over $P \in [0.2, 1.0]$. Here we extend the sweep to $P{=}5\%$ on SmolLM3-3B-Base (WIKI) under the identical protocol, evaluating the DLN pool with the LAPE budget matched to ours at every $P$. As Table~\ref{tab:psweep} shows, our selectivity remains an order of magnitude above LAPE ($9$--$14\times$) at every extended operating point.

The mild decline of the raw on-target damage ($0.40 \rightarrow 0.29$) is a composition effect rather than a weakening of the method: higher $P$ admits additional three-language MLN clusters with small per-cluster damage ($13 \rightarrow 33$ clusters), which dilutes the unweighted cluster mean in the selectivity numerator. Within the SLN stratum, whose cluster count is fixed at seven, on-target damage instead rises monotonically ($1.47 \rightarrow 1.64$) while off-target damage stays below $0.02$ nats. The percentile choice is therefore not load-bearing anywhere in the extended range.

\subsection{Histogram Bin-count Ablation}
\label{app:bins}
\begin{table}[!h]
    \centering
    \small
    \setlength{\tabcolsep}{5pt}
    \begin{tabular}{r r r r r}
    \toprule
    \#bins & $\tau$ & $|\mathcal{N}_{\mathrm{SLN}}|$ & Jaccard & time (s) \\
    \midrule
    10 & 0.4231 & 3,116 & 0.919 & 0.8 \\
    50 & 0.4015 & 2,983 & 0.996 & 1.7 \\
    100 & 0.4006 & 2,975 & 0.998 & 3.4 \\
    200 & 0.4003 & 2,974 & 0.999 & 6.1 \\
    400 & 0.4001 & 2,972 & 1.000 & 11.6 \\
    800 & 0.3998 & 2,973 & 0.999 & 23.5 \\
    \bottomrule
    \end{tabular}
    \caption{Bin-count ablation on Llama-3.1-8B ($P{=}1\%$). Jaccard is measured against the paper's $400$-bin selection; time is the CPU cost of the overlap computation and selection. Target-language assignments agree at $100\%$ for every bin count.}
    \label{tab:bins}
\end{table}

Our identifier approximates each per-language density with a $400$-bin adaptive histogram (Section~\ref{subsec:experiment_settings}). To probe this choice, we rebuild the histogram cache at $800$ bins on Llama-3.1-8B and derive every coarser resolution by aggregating adjacent bins, which is exact because the per-neuron bin edges nest; each resolution is then passed through the identical selection at $P{=}1\%$. As Table~\ref{tab:bins} shows, the selection is essentially invariant from $50$ to $800$ bins, a $16\times$ range. The threshold $\tau$ moves by less than $0.002$, the SLN pool stays within Jaccard $0.996$ of the paper's, and every selected neuron keeps the same target-language assignment at every resolution, including $10$ bins. Only at $10$ bins does the pool itself drift (Jaccard $0.92$), as coarse bins begin to blur the near-zero negative-region structure that distinguishes English neurons. Since the bin count affects only the post-forward overlap computation, which takes seconds on CPU, the resolution can be reduced substantially at no cost to the selection.

\section{Robustness of the Identifier Design}
\begin{table}[!h]
\centering
\small
\setlength{\tabcolsep}{4pt}
\resizebox{\linewidth}{!}{
\begin{tabular}{lcccc}
\toprule
Variant & $|\mathcal{N}|$ & Jac. & $\Delta\mathrm{NLL}$ & Sel. \\
        &                 &      & (W\,/\,F)            & (W\,/\,F) \\
\midrule
\multicolumn{5}{c}{\textit{Linkage (overlap statistic)}} \\
overlap, single (ours) & 2{,}972 & 1.00 & 1.20 / 1.69 & 198 / 408 \\
overlap, average       & 2{,}939 & 0.99 & 1.14 / 1.60 & 198 / 403 \\
overlap, complete      & 2{,}848 & 0.96 & 0.97 / 1.37 & 179 / 374 \\
\midrule
\multicolumn{5}{c}{\textit{Bounded statistic (single linkage)}} \\
Jensen--Shannon        & 2{,}999 & 0.95 & 1.27 / 1.83 & 202 / 424 \\
squared Hellinger      & 3{,}066 & 0.92 & 1.29 / 1.90 & 207 / 445 \\
Kolmogorov--Smirnov    & 2{,}970 & 0.99 & 1.17 / 1.64 & 201 / 400 \\
\midrule
\multicolumn{5}{c}{\textit{Scale-bearing statistic (single linkage)}} \\
Wasserstein-1          & 5{,}603 & 0.47 & 0.96 / 1.31 & \phantom{0}81 / 152 \\
energy distance        & 4{,}797 & 0.58 & 0.99 / 1.31 & 120 / 234 \\
\quad W1, top-$|\mathcal{N}|$      & 2{,}972 & --- & 0.62 / 0.85 & 117 / 212 \\
\quad energy, top-$|\mathcal{N}|$  & 2{,}972 & --- & 0.77 / 1.05 & 156 / 302 \\
\bottomrule
\end{tabular}
}
\caption{Effect of the pairwise statistic and linkage rule on the selected pool (Llama-3.1-8B, $P=1\%$). Jac.\ is the Jaccard overlap with our overlap$+$single set; $\Delta\mathrm{NLL}$ is the on-target damage and Sel.\ the selectivity, each on WIKI (W) and FLORES (F). The last two rows truncate the scale-bearing pools to our pool size by top-ranked score (neuron-matched control).}
\label{tab:metric-linkage}
\end{table}

\label{app:metric-linkage}

We examine three design choices of our identifier on Llama-3.1-8B at $P{=}1\%$. These are the pairwise overlap coefficient $\mathbf{S}_j$ (Section~\ref{subsec:cluster}), the single-linkage bipartitioning, and the fixed cluster count $K{=}2$. For a general pairwise distance, we take the most-separated bipartition and threshold at the top-$P$ percentile. Table~\ref{tab:metric-linkage} reports the pool size $|\mathcal{N}|$, its Jaccard overlap with our set, and the on-target $\Delta\mathrm{NLL}$ and selectivity on WIKI\,/\,FLORES.

\paragraph{Bounded statistics and linkage agree.}
Every bounded, shape-sensitive statistic selects essentially the same neuron set. Jensen--Shannon, squared Hellinger, and Kolmogorov--Smirnov stay within $3\%$ of our pool size, with Jaccard $0.92$--$0.99$ and selectivity within $\sim\!10\%$ on both corpora. They also preserve the multi-language structure, with Jensen--Shannon recovering $650$ MLNs against our $654$ under the same leading coalitions. The linkage rule is equally immaterial (Jaccard $0.96$--$0.99$ across single/average/complete), and we keep single linkage because it is exactly the stated max-overlap criterion.

\paragraph{Scale-bearing metrics reward magnitude, not discriminability.}
Wasserstein-1 and the energy distance instead inflate the pool ($1.6$--$1.9\times$), fall to Jaccard $0.47$--$0.58$, roughly halve selectivity, and distort the MLN structure. This is not merely over-selection but a property of the ranking itself. Truncating each pool to our size by top-ranked score makes them worse rather than better ($0.62$ vs.\ $0.96$ on-target $\Delta\mathrm{NLL}$ for W1 on WIKI). Their ranking places the largest-variance neurons first.

\paragraph{Why the overlap coefficient.}
Among these interchangeable statistics, we adopt the overlap coefficient for interpretability and cost. It equals one minus the total variation distance, $\mathbf{S}_j[k,k']=1-\mathrm{TV}(p_j^k, p_j^{k'})$, so $\tau$ reads directly as a bound on shared probability mass, and it is the cheapest to compute over all $458$k neurons ($10$\,s on CPU, against $96$\,s for Jensen--Shannon).

\paragraph{The $K{=}2$ bipartition is a selection decision, not a bimodality assumption.}
For every neuron we recompute the selection statistic inside the larger reference cluster $A_j^c$ of its most-separated bipartition, so a genuinely three-group neuron clears $\tau$ on this residual split as well. The empirical structure is overwhelmingly two-group. The median residual-split overlap is $0.78$ among selected neurons and $0.92$ among rejected ones, nearly twice the threshold $\tau{=}0.40$, and only $50$ of the $3{,}626$ selected neurons ($1.4\%$) admit a threshold-clearing residual split, with linguistically coherent nestings such as $\{$\texttt{zh}, \texttt{ja}$\}$ then $\{$\texttt{en}$\}$. Moreover, none of the $455{,}126$ rejected neurons admits a three-group structure, which follows from the criterion itself, since three mutually separated groups already provide a bipartition whose cross-cluster overlap clears $\tau$. Such neurons are selected through their most-separated split and labeled by its minority side. A natural refinement is to choose the number of clusters by within-cluster cohesion at additional computational cost, which we leave to future work.

\section{Displacement-matched Random Control}
\label{app:displacement-control}

\begin{table}[t]
    \centering
    \small
    \setlength{\tabcolsep}{4pt}
    \begin{tabular}{l r rrr r}
    \toprule
    & & \multicolumn{3}{c}{on-target $\Delta\mathrm{NLL}_k$} & \\
    \cmidrule(lr){3-5}
    $k$ & $d$ ($\sigma$) & SLN & Rand & Rand$\,{\times}\frac{1}{2}$ & Ratio \\
    \midrule
    \texttt{en} & 1.61 & 1.466 & 0.010 & 0.005 & $148\times$ \\
    \texttt{zh} & 1.85 & 0.009 & 0.001 & 0.000 & $15\times$ \\
    \texttt{fr} & 2.12 & 1.208 & 0.021 & 0.017 & $58\times$ \\
    \texttt{es} & 2.05 & 1.843 & 0.014 & 0.006 & $131\times$ \\
    \texttt{vi} & 1.91 & 1.662 & 0.007 & 0.006 & $242\times$ \\
    \texttt{id} & 2.04 & 0.605 & 0.028 & 0.011 & $21\times$ \\
    \texttt{ja} & 2.17 & 1.596 & 0.010 & 0.001 & $155\times$ \\
    \midrule
    mean & --- & 1.198 & 0.013 & 0.006 & $92\times$ \\
    \bottomrule
    \end{tabular}
    \caption{Displacement-matched random control on Llama-3.1-8B (WIKI, $P{=}1\%$). $d$ is the normalized displacement of the matched SLN set; Rand patches random layer- and count-matched neurons by the same $d$, Rand$\,{\times}\frac{1}{2}$ by half. Ratio is SLN over Rand on-target damage. The mean ratio is the ratio of the mean damages.}
    \label{tab:displacement_control}
\end{table}

For each SLN identified for target language $k$, we measure its normalized displacement under mean-patching, i.e., the gap between its mean activation on the target language and its mean over the non-target languages, in units of its pooled standard deviation. We then draw random neurons outside every selected pool and the LAPE pool, matched to the SLN set in layer and count, and shift each away from its own target-language mean by the same normalized amount and direction (3 seeds; we report the mean). A \emph{half-displacement} condition repeats this at half the magnitude as a dose--response probe. Table~\ref{tab:displacement_control} reports per-language results on Llama-3.1-8B (WIKI, $P{=}1\%$).

Three observations follow. (i) The same displacement applied at random locations yields ${\sim}1\%$ of the SLN effect, so displacement magnitude does not mechanically produce the reported damage. (ii) The control has no language selectivity: its on-target and off-target damage are identical ($0.013$ vs.\ $0.013$), i.e., random displacement acts as diffuse noise, while the SLN intervention concentrates its effect on the target language with off-target damage ($0.006$) below the control's noise floor. (iii) Halving the displacement leaves the control equally inert ($0.006$), ruling out a dose--response account. The smallest ratio, Chinese, is small for both interventions (only $27$ Chinese SLNs at $P{=}1\%$); as in Section~\ref{subsec:q1_identification}, the Chinese signal resides in the $\{$\texttt{zh}, \texttt{ja}$\}$ MLN cluster. All three observations replicate on SmolLM3-3B-Base: control $0.010$ vs.\ SLN $1.395$ on-target, per-language ratios $23$--$378\times$, and no dose response ($0.005$ at half displacement).

\section{Downstream Evaluation}

\subsection{Experimental Settings}
\label{app:downstream_setup}

\paragraph{Belebele.} Each question is scored zero-shot. We build one prompt per answer option,
\begin{quote}
\ttfamily
\begin{tabular}{@{}l@{}}
\{passage\}\\
Q: \{question\}\\
A: \{option\}
\end{tabular}
\end{quote}
\noindent and take the summed log-probability of the \texttt{\{option\}} tokens as its score, predicting the highest-scoring option. 

\paragraph{MGSM.} Each problem is prompted with the $8$ chain-of-thought exemplars from the dataset's own training split for that language, so the exemplars, their reasoning, and the question prefixes are all in the evaluation language. The prompt is
\begin{quote}
\ttfamily
\begin{tabular}{@{}l@{}}
\{q$_1$\}\\
\{a$_1$\}\\[4pt]
$\vdots$\\[4pt]
\{q$_8$\}\\
\{a$_8$\}\\[4pt]
\{test question\}\\
\{answer prefix\}
\end{tabular}
\end{quote}
\noindent where each question sits on the line above its answer and the blocks are separated by a blank line, so the trailing answer prefix leaves the model to produce the reasoning. Table~\ref{tab:mgsm_prefixes} lists the prefixes per language. We decode greedily for at most $512$ new tokens, stopping at a blank line or at a new question prefix, and take the last number in the generation as the predicted answer, matched exactly against the gold value.

\begin{table}[h]
\centering
\small
\begin{tabular}{lll}
\toprule
& \textbf{Question prefix} & \textbf{Answer prefix} \\
\midrule
\texttt{en} & Question: & Step-by-Step Answer: \\
\texttt{zh} & \begin{CJK}{UTF8}{gbsn}问题：\end{CJK} & 逐步解答： \\
\texttt{fr} & Question~: & Réponse étape par étape~: \\
\texttt{es} & Pregunta: & Respuesta paso a paso: \\
\texttt{ja} & \begin{CJK}{UTF8}{min}問題：\end{CJK} & ステップごとの答え： \\
\bottomrule
\end{tabular}
\caption{MGSM prompt prefixes for the five languages MGSM covers.}
\label{tab:mgsm_prefixes}
\end{table}

\subsection{Per-language Downstream Results}
\label{app:downstream}

\begin{table}[t]
\centering
\scriptsize
\setlength{\tabcolsep}{3.5pt}
\resizebox{\linewidth}{!}{
\begin{tabular}{ll rrrrrrr r}
\toprule
& \textbf{Method} & \texttt{en} & \texttt{zh} & \texttt{fr} & \texttt{es} & \texttt{vi} & \texttt{id} & \texttt{ja} & \textbf{Avg} \\
\midrule
\multicolumn{10}{c}{\textbf{Llama-3.1-8B}} \\
\midrule
\multirow{3}{*}{$|\mathcal{N}_{\{k\}}|$}
& \lape       & 61 & 715 & 822 & 619 & 442 & 1057 & 871 & -- \\
& \ternarylsn & 28 & 107 & 481 & 318 & 105 & 540 & 235 & -- \\
& SLN         & 188 & 27 & 780 & 616 & 199 & 915 & 247 & -- \\
\cmidrule(lr){1-10}
\multirow{3}{*}{Belebele}
& \lape       & 0.33 & \textbf{3.33} & 0.67 & 1.44 & 4.56 & \textbf{2.67} & 6.22 & 2.75 \\
& \ternarylsn & 0.00 & 1.56 & 0.56 & 0.56 & 3.44 & 0.89 & 1.56 & 1.22 \\
& SLN         & \textbf{6.56} & 0.11 & \textbf{5.00} & \textbf{7.11} & \textbf{6.78} & 2.44 & \textbf{6.56} & \textbf{4.94} \\
\cmidrule(lr){1-10}
\multirow{3}{*}{MGSM}
& \lape       & 0.40 & \textbf{8.00} & 0.80 & 2.00 & -- & -- & \textbf{23.60} & 6.96 \\
& \ternarylsn & $-$0.80 & 3.20 & 0.80 & 4.80 & -- & -- & 14.00 & 4.40 \\
& SLN         & \textbf{12.00} & $-$0.40 & \textbf{11.60} & \textbf{8.80} & -- & -- & 15.20 & \textbf{9.44} \\
\midrule
\multicolumn{10}{c}{\textbf{SmolLM3-3B-Base}} \\
\midrule
\multirow{3}{*}{$|\mathcal{N}_{\{k\}}|$}
& \lape       & 45 & 657 & 539 & 431 & 547 & 916 & 827 & -- \\
& \ternarylsn & 16 & 119 & 296 & 207 & 201 & 375 & 229 & -- \\
& SLN         & 152 & 44 & 475 & 425 & 357 & 566 & 235 & -- \\
\cmidrule(lr){1-10}
\multirow{3}{*}{Belebele}
& \lape       & 0.11 & \textbf{1.78} & 1.56 & 1.00 & 6.89 & 2.00 & 3.89 & 2.46 \\
& \ternarylsn & 0.00 & $-$0.33 & 0.78 & 0.56 & 5.44 & 1.33 & 1.00 & 1.25 \\
& SLN         & \textbf{3.22} & $-$0.22 & \textbf{7.89} & \textbf{9.00} & \textbf{8.33} & \textbf{4.78} & \textbf{5.89} & \textbf{5.56} \\
\cmidrule(lr){1-10}
\multirow{3}{*}{MGSM}
& \lape       & 3.60 & 0.00 & 0.00 & 1.20 & -- & -- & 13.20 & 3.60 \\
& \ternarylsn & 0.40 & \textbf{1.60} & $-$0.40 & 3.60 & -- & -- & $-$1.60 & 0.72 \\
& SLN         & \textbf{10.00} & 0.40 & \textbf{5.20} & \textbf{9.20} & -- & -- & \textbf{14.00} & \textbf{7.76} \\
\bottomrule
\end{tabular}
}
\caption{Per-language on-target accuracy drop under the mean-patch intervention, expanding Table~\ref{tab:downstream}. The $|\mathcal{N}_{\{k\}}|$ block gives the number of neurons each identifier assigns to target language $k$; these pools are shared between the two tasks, so the MGSM totals in Table~\ref{tab:downstream} sum only over the five languages MGSM covers. \textbf{Bold} denotes the largest drop per column.}
\label{tab:downstream_per_lang}
\end{table}

Table~\ref{tab:downstream_per_lang} breaks Table~\ref{tab:downstream} down by target language. English separates the methods most sharply. Both baselines assign English a small subset, $61$ and $28$ neurons on Llama-3.1-8B and $45$ and $16$ on SmolLM3-3B-Base, and patching it costs at most $0.33$ points on Belebele and $3.60$ on MGSM, whereas our English SLNs remove $6.56$ and $12.00$ points on Llama-3.1-8B and $3.22$ and $10.00$ on SmolLM3-3B-Base. This reproduces on task accuracy the English gap reported for $\Delta\mathrm{NLL}$ in Section~\ref{subsec:q1_identification}, and it is the downstream counterpart of the negative-region analysis in Section~\ref{sec:lape_vs_sln}: $63$--$72\%$ of the English SLNs patched here have a negative target-language mean and therefore lie in the region that positive-rate metrics discard by construction.

Chinese is the one target where our SLNs fall below \lape, at $0.11$ versus $3.33$ on Belebele and $-0.40$ versus $8.00$ on MGSM for Llama-3.1-8B. This mirrors the $\Delta\mathrm{NLL}$ exception in Section~\ref{subsec:q1_identification} and follows from the same property of our identifier: almost all Chinese language neurons are assigned to the $\{$\texttt{zh}, \texttt{ja}$\}$ multi-language cluster, leaving $27$ neurons on Llama-3.1-8B and $44$ on SmolLM3-3B-Base as single-language, so MLNs account for $93.3\%$ and $88.1\%$ of the Chinese language neurons (Table~\ref{tab:mln_share_p1}). The Chinese SLN column therefore patches these $27$ and $44$ neurons, not the pool that carries the Chinese effect; that pool is the MLN cluster, where mean-patching $\mathcal{N}_{\{\texttt{zh},\texttt{ja}\}}$ yields $\Delta\mathrm{NLL}_\texttt{zh} = 1.24$ on WIKI and $1.12$ on FLORES.

Outside Chinese, a baseline exceeds our raw on-target drop in $2$ of the $20$ remaining (model, task, language) cells: Indonesian on Belebele with Llama-3.1-8B ($2.67$ versus $2.44$) and Japanese on MGSM with Llama-3.1-8B ($23.60$ versus $15.20$). In both, \lape\ patches a larger pool, $1057$ against $915$ neurons and $871$ against $247$, and our per-neuron drop is still larger, by $1.06\times$ and $2.27\times$. The averages in Table~\ref{tab:downstream} therefore do not rest on a few target languages.

\begin{table}[t]
\centering
\scriptsize
\setlength{\tabcolsep}{4pt}
\resizebox{\linewidth}{!}{
\begin{tabular}{ll rrr rrr}
\toprule
& & \multicolumn{3}{c}{\textbf{on-target} ($k^\prime{=}k$)} & \multicolumn{3}{c}{\textbf{off-target} ($k^\prime{\neq}k$)} \\
\cmidrule(lr){3-5} \cmidrule(lr){6-8}
& \textbf{Method} & $b$ & $c$ & $p$ & $b$ & $c$ & $p$ \\
\midrule
\multicolumn{8}{c}{\textbf{Llama-3.1-8B}} \\
\midrule
\multirow{3}{*}{Belebele}
& \lape       & 392 & 219 & $2\!\times\!10^{-12}$ & 409 & 381 & 0.34 \\
& \ternarylsn & 282 & 205 & $6\!\times\!10^{-4}$  & 109 & 118 & 0.60 \\
& SLN         & 612 & 301 & $4\!\times\!10^{-25}$ & 153 & 161 & 0.69 \\
\cmidrule(lr){1-8}
\multirow{3}{*}{MGSM}
& \lape       & 162 & 75 & $2\!\times\!10^{-8}$  & 297 & 253 & 0.07 \\
& \ternarylsn & 113 & 58 & $3\!\times\!10^{-5}$  & 90  & 75  & 0.28 \\
& SLN         & 212 & 94 & $1\!\times\!10^{-11}$ & 124 & 115 & 0.61 \\
\midrule
\multicolumn{8}{c}{\textbf{SmolLM3-3B-Base}} \\
\midrule
\multirow{3}{*}{Belebele}
& \lape       & 402 & 247 & $1\!\times\!10^{-9}$  & 510 & 458 & 0.10 \\
& \ternarylsn & 282 & 203 & $4\!\times\!10^{-4}$  & 113 & 110 & 0.89 \\
& SLN         & 685 & 335 & $3\!\times\!10^{-28}$ & 228 & 173 & \underline{0.007} \\
\cmidrule(lr){1-8}
\multirow{3}{*}{MGSM}
& \lape       & 127 & 82 & $2\!\times\!10^{-3}$ & 263 & 260 & 0.93 \\
& \ternarylsn & 79  & 70 & \underline{0.51}     & 70  & 102 & \underline{0.02} \\
& SLN         & 176 & 79 & $1\!\times\!10^{-9}$ & 111 & 131 & 0.22 \\
\bottomrule
\end{tabular}
}
\caption{Two-sided exact McNemar tests on the item-level outcomes behind Table~\ref{tab:downstream}, pooled over target languages. $b$ counts items answered correctly by the unintervened model and incorrectly after the intervention, and $c$ counts the reverse, so $b > c$ indicates a drop. \underline{Underline} marks the three entries discussed in the text.}
\label{tab:downstream_mcnemar}
\end{table}

Because the accuracy differences in Table~\ref{tab:downstream} are small in absolute terms, we also test them at the item level. Table~\ref{tab:downstream_mcnemar} reports two-sided exact McNemar tests on the discordant pairs between the unintervened and intervened runs, pooled over target languages. Every on-target drop reaches $p < 0.01$ except \ternarylsn\ on MGSM with SmolLM3-3B-Base ($p = 0.51$), whose $0.72$-point entry in Table~\ref{tab:downstream} is therefore not distinguishable from no effect. Our SLNs give the smallest $p$-values in all four settings, at $p \le 4\!\times\!10^{-25}$ on Belebele and $p \le 1\!\times\!10^{-9}$ on MGSM.

Ten of the twelve off-target entries are not significant at the $0.05$ level. The two exceptions are our SLNs on Belebele with SmolLM3-3B-Base ($p = 0.007$), whose magnitude is $0.15$ accuracy points, and \ternarylsn\ on MGSM with SmolLM3-3B-Base ($p = 0.02$ with $c > b$, that is, off-target accuracy rises). Both are likely overstated: unlike the on-target test, where each target language contributes a disjoint item set, the off-target test reuses the same evaluation items under the six or four other target-language interventions, so the pooled discordant pairs are not independent and the test is anti-conservative.

\section{Concrete Examples of Activation Distribution}

\begin{figure*}[t]
  \centering
    \includegraphics[width=\linewidth]{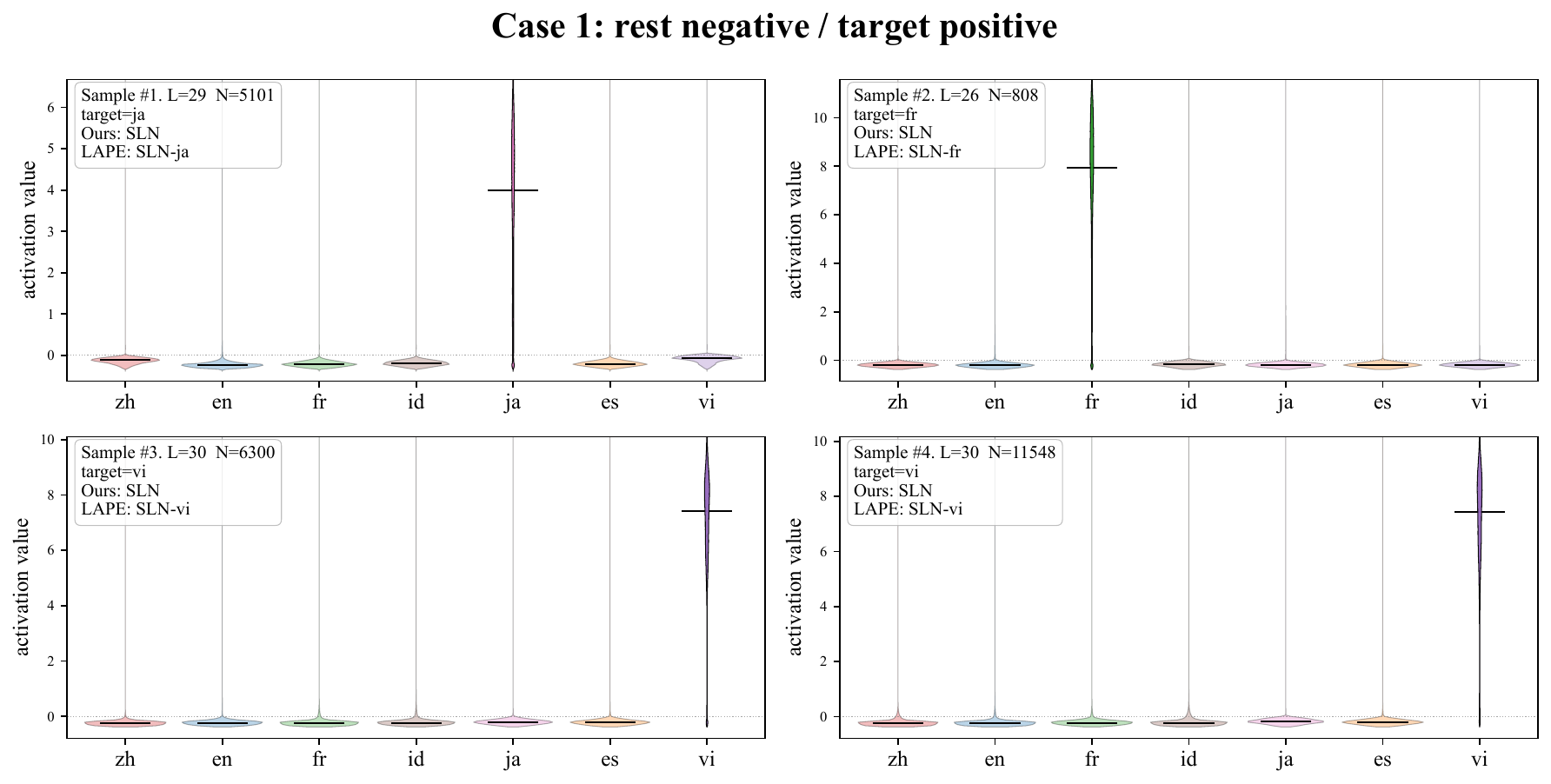}
  \caption{Four SLN samples for case 1 (rest negative / target positive). Both identifiers agree on the target-language SLN; inset box reports $(L, N, \text{target language}, \text{Ours / LAPE verdict})$.}
  \label{fig:appendix-case1-examples}
\end{figure*}

\begin{figure*}[t]
  \centering
    \includegraphics[width=\linewidth]{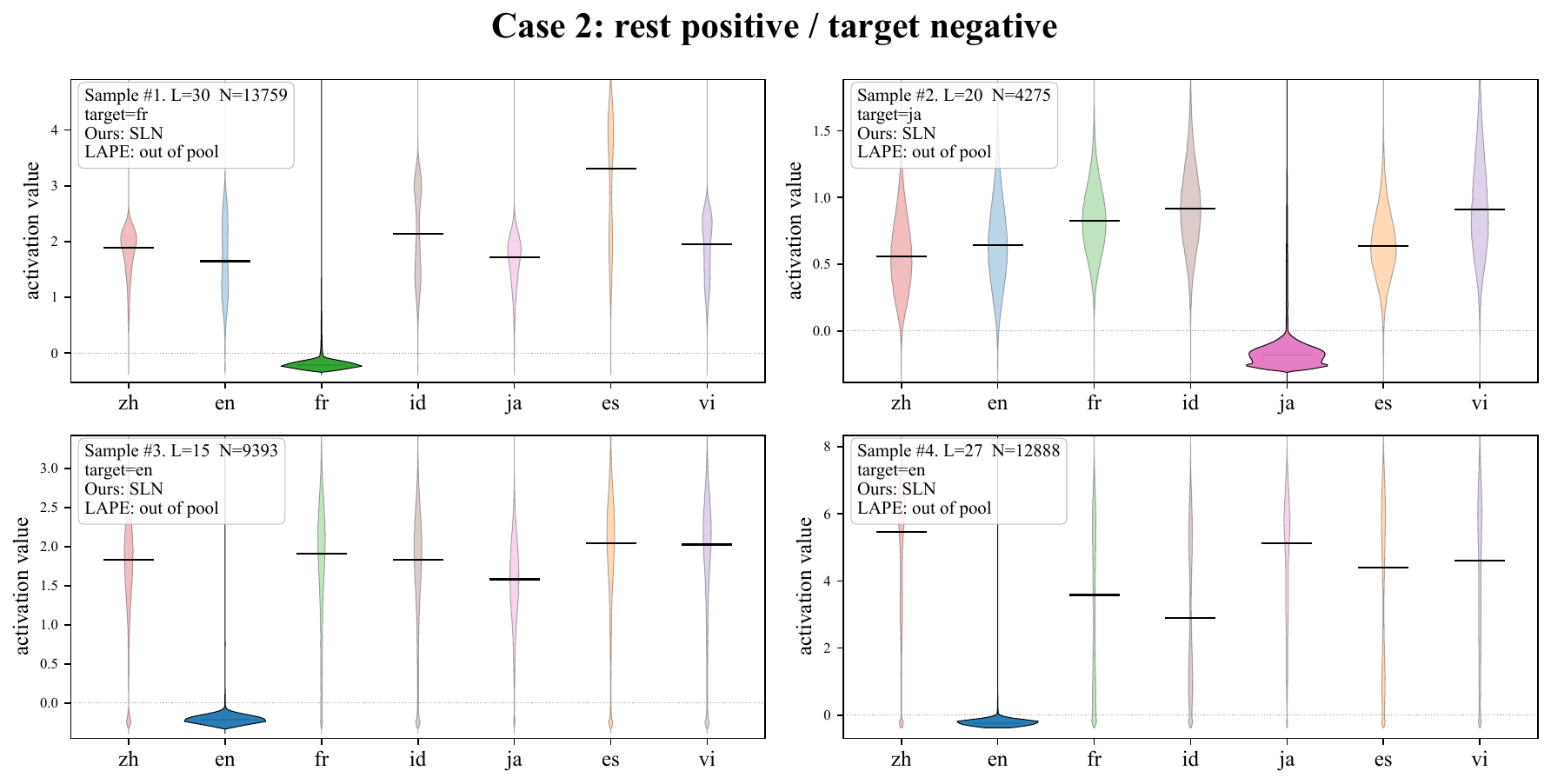}
  \caption{Four SLN samples for case 2 (rest positive / target negative). SLN assigns the target-language SLN while LAPE misses --- the sign reversal leaves the normalized activation-rate entropy near-uniform; inset box reports $(L, N, \text{target language}, \text{Ours / LAPE verdict})$}
  \label{fig:appendix-case2-examples}
\end{figure*}

\begin{figure*}[t]
  \centering
    \includegraphics[width=\linewidth]{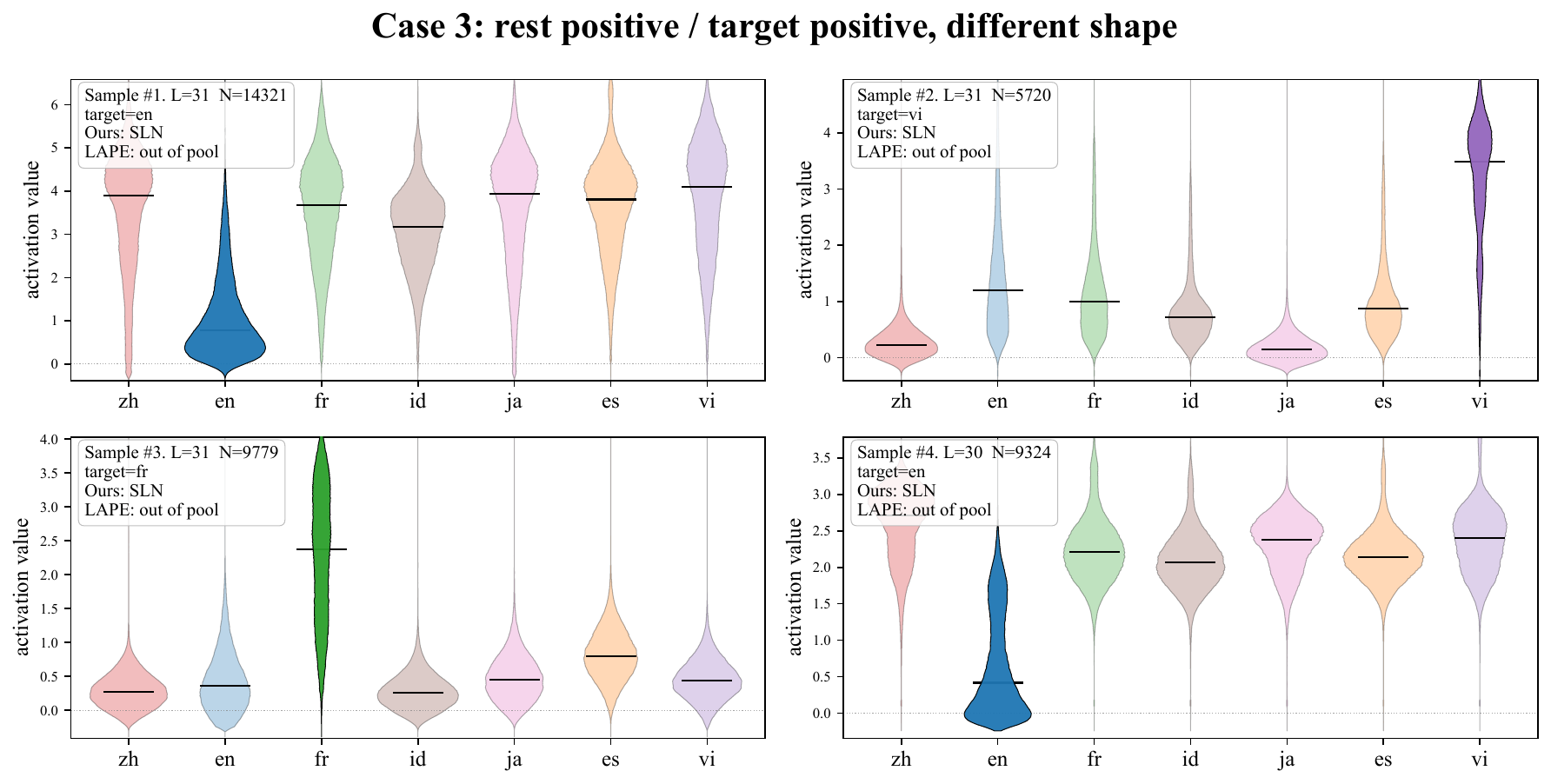}
  \caption{Four SLN samples for case 3 (rest positive / target positive, different shape). SLN assigns the target-language SLN while LAPE misses --- all languages clear $a>0$, so the activation-rate statistic carries no separating signal; inset box reports $(L, N, \text{target language}, \text{Ours / LAPE verdict})$.}
  \label{fig:appendix-case3-examples}
\end{figure*}

\begin{figure*}[t]
  \centering
    \includegraphics[width=\linewidth]{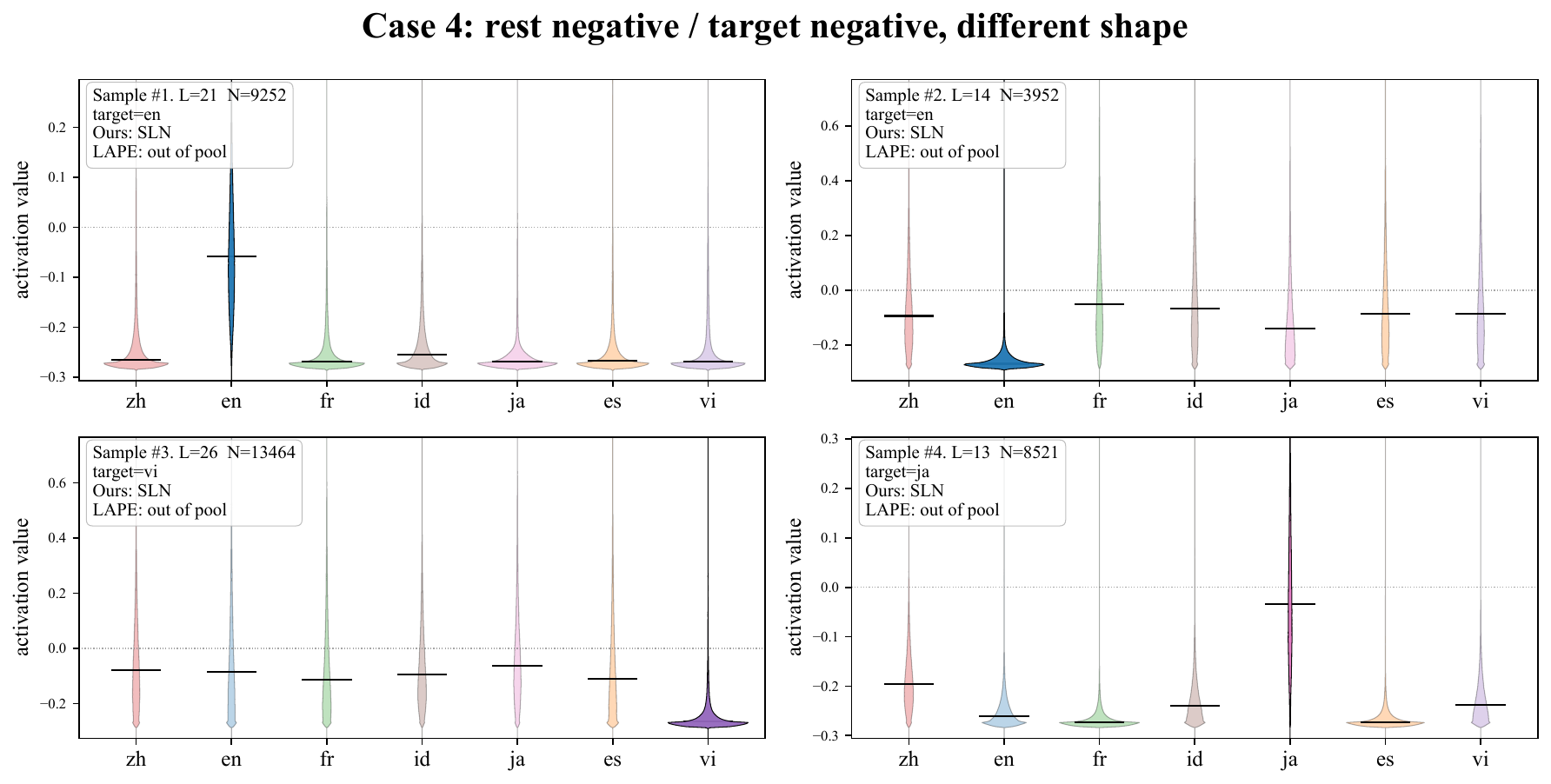}
  \caption{Four SLN samples for case 4 (rest negative / target negative, different shape). SLN assigns the target-language SLN while LAPE excludes these neurons from its pool --- no language clears the max-rate threshold $\tau$; inset box reports $(L, N, \text{target language}, \text{Ours / LAPE verdict})$}
  \label{fig:appendix-case4-examples}
\end{figure*}

In this section, we illustrate the four representative activation patterns presented in Figure~\ref{fig:representative_cases}. Figures~\ref{fig:appendix-case1-examples}--\ref{fig:appendix-case4-examples} display four SLN samples corresponding to each case as violin plots across the seven calibration languages, with each inset indicating the neuron's location $(L,N)$, the target language, and whether it is identified as SLN and/or LAPE.

Case~1 (Figure~\ref{fig:appendix-case1-examples}) is the canonical form in which only the target language activates in the positive region while the remaining languages are distributed in the negative region, and both identifiers identify the same neuron as a target-language SLN. Cases~2--4(Figure~\ref{fig:appendix-case2-examples}--\ref{fig:appendix-case4-examples}) are all SLN-only samples that LAPE fails to identify: case~2 is a sign-inverted form in which the target language lies in the negative region and the remaining languages lie in the positive region; case~3 is a form in which all languages reside in the positive region but the target language is separated only by magnitude; and case~4 is a form in which all languages are distributed in the negative region but only the target language is separated by magnitude.

\clearpage

\clearpage

\section{Detailed Neuron Population}
\subsection{Language Population in SLN, MLN} \label{appendix: MLN-analysis}

\begin{table}[h]
\centering
\scriptsize
\renewcommand{\arraystretch}{0.95}
\setlength{\tabcolsep}{4pt}
\begin{tabular}{l l r r r}
\toprule
Model & Language & $|\mathrm{SLN}|$ & $|\mathrm{MLN}|$ & Ratio \\
\midrule
Llama-3.1-8B & \texttt{zh} & 27 & 376 & \textbf{93.3\%} \\
& \texttt{ja} & 247 & 371 & \textbf{60.0\%} \\
& \texttt{vi} & 199 & 148 & 42.7\% \\
& \texttt{en} & 188 & 76 & 28.8\% \\
& \texttt{es} & 616 & 220 & 26.3\% \\
& \texttt{fr} & 780 & 198 & 20.2\% \\
& \texttt{id} & 915 & 84 & 8.4\% \\
\midrule
SmolLM3-3B & \texttt{zh} & 44 & 325 & \textbf{88.1\%} \\
& \texttt{ja} & 235 & 337 & \textbf{58.9\%} \\
& \texttt{en} & 152 & 75 & 33.0\% \\
& \texttt{vi} & 357 & 161 & 31.1\% \\
& \texttt{fr} & 475 & 133 & 21.9\% \\
& \texttt{es} & 425 & 119 & 21.9\% \\
& \texttt{id} & 566 & 76 & 11.8\% \\
\midrule
Qwen3-4B & \texttt{zh} & 113 & 567 & \textbf{83.4\%} \\
& \texttt{en} & 120 & 214 & \textbf{64.1\%} \\
& \texttt{ja} & 663 & 546 & 45.2\% \\
& \texttt{es} & 534 & 341 & 39.0\% \\
& \texttt{vi} & 287 & 181 & 38.7\% \\
& \texttt{fr} & 651 & 331 & 33.7\% \\
& \texttt{id} & 690 & 196 & 22.1\% \\
\bottomrule
\end{tabular}
\caption{Per-language counts of SLNs and MLNs, on Llama-3.1-8B, SmolLM3-3B-Base, and Qwen3-4B-Base. $|\cdot|$ denote the number of neurons associated with each language. Ratio $= \frac{|\mathrm{MLN}|}{|\mathrm{SLN}|+|\mathrm{MLN}|}$ gives the share of MLNs among all language neurons.}
\label{tab:mln_share_p1}
\end{table}

\begin{figure}[t]
  \centering
    \includegraphics[width=\linewidth]{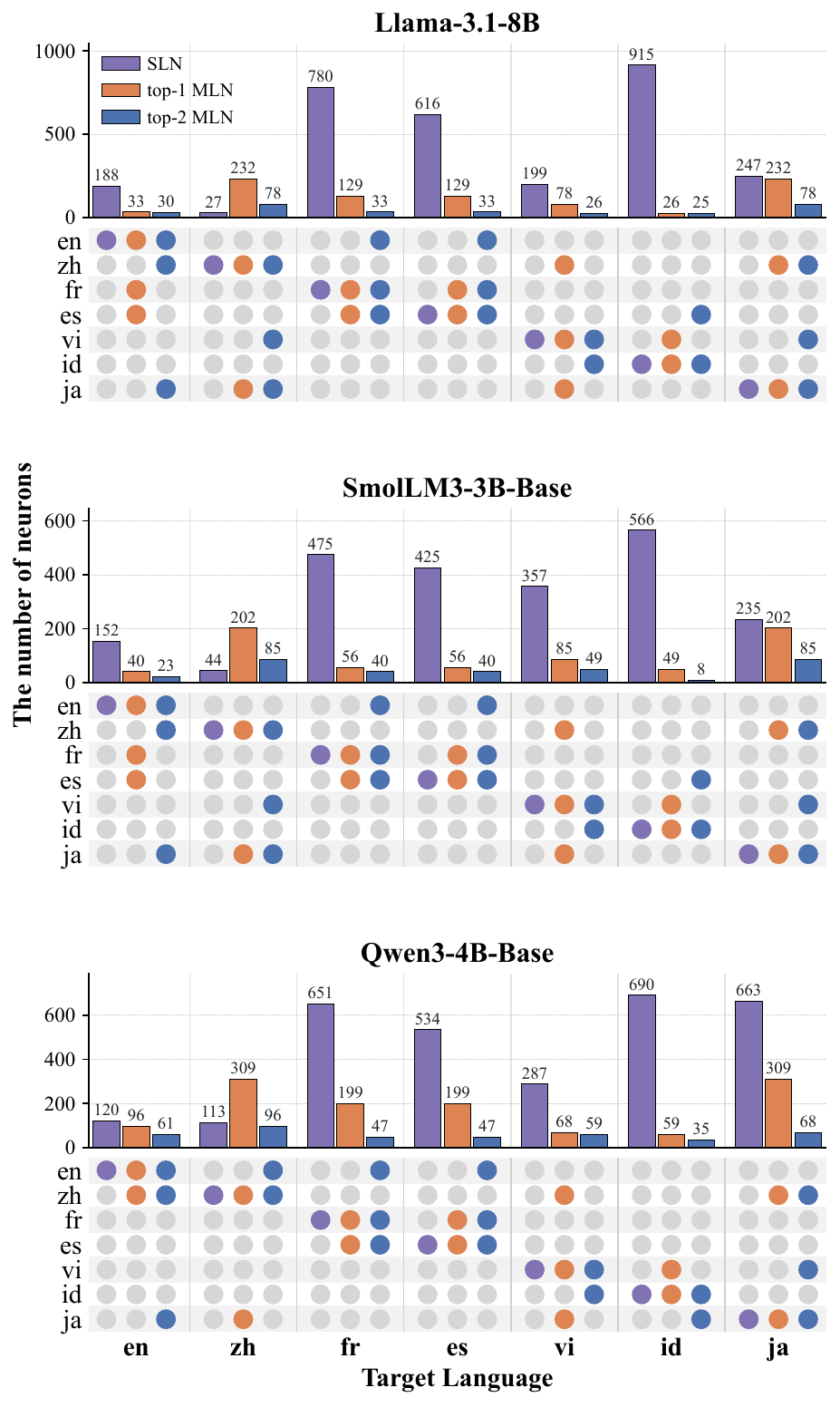}
  \caption{Language distribution of SLN and MLN neurons across different languages in three language models. The top-$n$ MLN denotes the MLN cluster with the $n$-th highest frequency among all MLN clusters that include the corresponding language.}
  \label{fig:appendix-upset-mln-whole}
\end{figure}

Figure~\ref{fig:appendix-upset-mln-whole} illustrates the number of SLNs and MLNs for each language across three models: Llama-3.1-8B, SmolLM3-3B-Base, and Qwen3-4B-Base. The ``top-$n$ MLN'' denotes the $n$-th most frequent cluster among the MLNs associated with a given language. The overall distributions of SLNs and MLNs exhibit similar patterns across all three models. For SLNs, \texttt{id} accounts for the highest proportion, whereas \texttt{zh} exhibits the lowest distribution in all models. Notably, for Chinese (\texttt{zh}), the proportion of MLNs is higher than that of SLNs across all models, with the $\{$\texttt{zh}, \texttt{ja}$\}$ cluster accounting for the largest share among the MLN populations. This indicates that Chinese and Japanese share the most information within these models. 

The distributions of the most frequent MLN clusters, such as $\{$\texttt{fr}, \texttt{es}$\}$, $\{$\texttt{vi}, \texttt{zh}, \texttt{ja}$\}$, and $\{$\texttt{id}, \texttt{vi}$\}$, are also consistent across the models. These results demonstrate that the proposed methodology is model-agnostic. Furthermore, this demonstrates that our proposed methodology enables the analysis of linguistic similarities and relational dynamics, such as those between Chinese and Japanese.

While the $\{$\texttt{en}, \texttt{fr}, \texttt{es}$\}$ cluster is the most frequent among English MLNs in the other two models, the $\{$\texttt{en}, \texttt{zh}$\}$ cluster is the most frequent in the Qwen3-4B-Base model. This discrepancy can be attributed to the Qwen3-4B-Base model being heavily specialized in Chinese compared to the others, treating it as a major language with a representation hierarchy comparable to English.

\subsection{Layer-wise Neuron Population}
In this section, we examine the layer-wise distributions and differences of the neurons identified by the previous works and our proposed approach. Similar to LAPE, both methodologies exhibit a U-shaped distribution, where the identified neurons are concentrated in the early and late layers. A key difference is that the neurons selected by our proposed methodology are more densely clustered in the extreme early and late layers compared to LAPE. Furthermore, MLNs are rarely observed in the initial layers but become increasingly prevalent in deeper layers. Detailed layer-wise statistics are provided in Tables~\ref{tab:neuron-population-llama-3.1-8b} and \ref{tab:neuron-population-smollm3-3b-base}.

\begin{table*}[t]
\centering
\small
\setlength{\tabcolsep}{5pt}
\begin{tabular}{rrrrrrrrr}
\toprule
Layer & LAPE & LSN & SLN & LRN & MLN & LAPE$\setminus$SLN & SLN$\setminus$LAPE & LAPE$\cap$SLN \\
\midrule
0 & 43 & 6 & 22 & 15 & 2 & 42 & 21 & 1 \\
1 & 2 & 3 & 11 & 5 & 2 & 2 & 11 & 0 \\
2 & 11 & 6 & 5 & 2 & 0 & 8 & 2 & 3 \\
3 & 90 & 40 & 91 & 106 & 17 & 56 & 57 & 34 \\
4 & 34 & 7 & 7 & 19 & 6 & 30 & 3 & 4 \\
5 & 19 & 4 & 7 & 14 & 3 & 16 & 4 & 3 \\
6 & 12 & 2 & 6 & 4 & 1 & 10 & 4 & 2 \\
7 & 5 & 0 & 4 & 3 & 0 & 5 & 4 & 0 \\
8 & 23 & 9 & 3 & 3 & 1 & 23 & 3 & 0 \\
9 & 36 & 8 & 8 & 23 & 11 & 34 & 6 & 2 \\
10 & 34 & 5 & 9 & 27 & 9 & 30 & 5 & 4 \\
11 & 21 & 5 & 5 & 12 & 2 & 20 & 4 & 1 \\
12 & 15 & 1 & 4 & 10 & 3 & 14 & 3 & 1 \\
13 & 8 & 3 & 11 & 13 & 1 & 7 & 10 & 1 \\
14 & 29 & 7 & 12 & 20 & 13 & 24 & 7 & 5 \\
15 & 38 & 13 & 14 & 23 & 12 & 34 & 10 & 4 \\
16 & 75 & 16 & 29 & 47 & 20 & 66 & 20 & 9 \\
17 & 152 & 28 & 36 & 92 & 22 & 132 & 16 & 20 \\
18 & 146 & 41 & 37 & 69 & 20 & 125 & 16 & 21 \\
19 & 138 & 36 & 39 & 47 & 12 & 111 & 12 & 27 \\
20 & 126 & 33 & 40 & 37 & 10 & 105 & 19 & 21 \\
21 & 162 & 46 & 56 & 37 & 12 & 132 & 26 & 30 \\
22 & 179 & 46 & 60 & 29 & 9 & 144 & 25 & 35 \\
23 & 112 & 35 & 64 & 39 & 9 & 85 & 37 & 27 \\
24 & 223 & 83 & 116 & 65 & 12 & 163 & 56 & 60 \\
25 & 212 & 69 & 111 & 72 & 18 & 165 & 64 & 47 \\
26 & 416 & 143 & 200 & 136 & 32 & 332 & 116 & 84 \\
27 & 604 & 262 & 342 & 195 & 37 & 458 & 196 & 146 \\
28 & 691 & 314 & 504 & 297 & 43 & 515 & 328 & 176 \\
29 & 406 & 233 & 473 & 344 & 61 & 304 & 371 & 102 \\
30 & 348 & 238 & 423 & 407 & 88 & 254 & 329 & 94 \\
31 & 177 & 72 & 223 & 504 & 166 & 152 & 198 & 25 \\
\midrule
Total & 4587 & 1814 & 2972 & 2716 & 654 & 3598 & 1983 & 989 \\
\bottomrule
\end{tabular}
\caption{Per-layer neuron-population counts for Llama-3.1-8B at $P\!=\!1\%$. Each cell is the number of neurons of the given population in that layer; set-difference and intersection columns are over (layer, neuron) pairs.}
\label{tab:neuron-population-llama-3.1-8b}
\end{table*}

\begin{table*}[t]
\centering
\small
\setlength{\tabcolsep}{5pt}
\begin{tabular}{rrrrrrrrr}
\toprule
Layer & LAPE & LSN & SLN & LRN & MLN & LAPE$\setminus$SLN & SLN$\setminus$LAPE & LAPE$\cap$SLN \\
\midrule
0 & 44 & 81 & 131 & 20 & 0 & 23 & 110 & 21 \\
1 & 8 & 5 & 5 & 0 & 0 & 5 & 2 & 3 \\
2 & 145 & 65 & 80 & 15 & 4 & 80 & 15 & 65 \\
3 & 51 & 19 & 14 & 8 & 1 & 44 & 7 & 7 \\
4 & 243 & 59 & 41 & 58 & 3 & 216 & 14 & 27 \\
5 & 9 & 1 & 17 & 3 & 1 & 7 & 15 & 2 \\
6 & 3 & 2 & 2 & 2 & 1 & 3 & 2 & 0 \\
7 & 19 & 5 & 4 & 4 & 3 & 19 & 4 & 0 \\
8 & 18 & 5 & 2 & 4 & 2 & 17 & 1 & 1 \\
9 & 14 & 2 & 0 & 3 & 0 & 14 & 0 & 0 \\
10 & 4 & 0 & 1 & 2 & 1 & 4 & 1 & 0 \\
11 & 6 & 1 & 1 & 1 & 2 & 6 & 1 & 0 \\
12 & 6 & 1 & 0 & 3 & 1 & 6 & 0 & 0 \\
13 & 2 & 0 & 3 & 5 & 0 & 2 & 3 & 0 \\
14 & 5 & 1 & 3 & 3 & 1 & 5 & 3 & 0 \\
15 & 16 & 0 & 6 & 13 & 3 & 13 & 3 & 3 \\
16 & 6 & 2 & 2 & 7 & 1 & 4 & 0 & 2 \\
17 & 34 & 9 & 8 & 18 & 6 & 29 & 3 & 5 \\
18 & 15 & 4 & 14 & 19 & 3 & 15 & 14 & 0 \\
19 & 56 & 15 & 14 & 43 & 25 & 49 & 7 & 7 \\
20 & 46 & 10 & 8 & 28 & 16 & 44 & 6 & 2 \\
21 & 39 & 11 & 18 & 41 & 8 & 33 & 12 & 6 \\
22 & 82 & 31 & 28 & 56 & 13 & 60 & 6 & 22 \\
23 & 103 & 33 & 36 & 74 & 26 & 82 & 15 & 21 \\
24 & 161 & 52 & 47 & 59 & 18 & 131 & 17 & 30 \\
25 & 150 & 43 & 37 & 63 & 18 & 128 & 15 & 22 \\
26 & 147 & 30 & 43 & 75 & 16 & 128 & 24 & 19 \\
27 & 208 & 64 & 67 & 80 & 24 & 173 & 32 & 35 \\
28 & 192 & 42 & 58 & 65 & 12 & 164 & 30 & 28 \\
29 & 225 & 56 & 78 & 65 & 11 & 198 & 51 & 27 \\
30 & 281 & 103 & 180 & 164 & 20 & 237 & 136 & 44 \\
31 & 253 & 106 & 225 & 166 & 18 & 210 & 182 & 43 \\
32 & 291 & 134 & 221 & 145 & 24 & 246 & 176 & 45 \\
33 & 296 & 144 & 249 & 171 & 24 & 228 & 181 & 68 \\
34 & 390 & 143 & 297 & 345 & 74 & 297 & 204 & 93 \\
35 & 394 & 164 & 314 & 520 & 150 & 293 & 213 & 101 \\
\midrule
Total & 3962 & 1443 & 2254 & 2348 & 530 & 3213 & 1505 & 749 \\
\bottomrule
\end{tabular}
\caption{Per-layer neuron-population counts for SmolLM3-3B-Base at $P\!=\!1\%$. Each cell is the number of neurons of the given population in that layer; set-difference and intersection columns are over (layer, neuron) pairs.}
\label{tab:neuron-population-smollm3-3b-base}
\end{table*}